\documentclass[10pt,twocolumn,letterpaper]{article}

\usepackage{cvpr}              
\definecolor{cvprblue}{rgb}{0.21,0.49,0.74}
\usepackage[pagebackref,breaklinks,colorlinks,allcolors=cvprblue]{hyperref}

\usepackage{graphicx}
\usepackage{booktabs}
\usepackage{multirow}
\usepackage[table]{xcolor}
\usepackage{placeins}
\usepackage{subcaption}
\usepackage{wrapfig}
\usepackage{float}
\usepackage{tcolorbox}      
\usepackage{listings}       
\usepackage{enumitem}       
\usepackage{caption}        

\definecolor{colorbest}{RGB}{252,187,161}
\definecolor{colorsecond}{RGB}{254,224,210}
\definecolor{colorthird}{RGB}{255,245,240}
\definecolor{dataconstruction}{RGB}{109,153,255}
\newcommand{\best}[0]{\cellcolor{colorbest} }
\newcommand{\second}[0]{\cellcolor{colorsecond}}
\newcommand{\third}[0]{\cellcolor{colorthird}}
\DeclareRobustCommand{\legendsquare}[1]{%
  \textcolor{#1}{\rule{2ex}{2ex}}%
}

\newcommand{\ourmethod}{PixelSmile}
\newcommand{\ourdataset}{FFE}
\newcommand{\ourbenchmark}{FFE-Bench}

\newcommand{\nbp}{Nano Banana Pro}
\newcommand{\gpt}{GPT-Image}
\newcommand{\seed}{Seedream}
\newcommand{\qwen}{Qwen-Edit}
\newcommand{\longcat}{LongCat}
\newcommand{\flux}{FLUX-Klein}
\newcommand{\kslider}{K-Slider}

\def\paperID{*****} 
\def\confName{CVPR}
\def\confYear{2026}

\title{\raisebox{-0.008\linewidth}{\includegraphics[width=0.6cm]{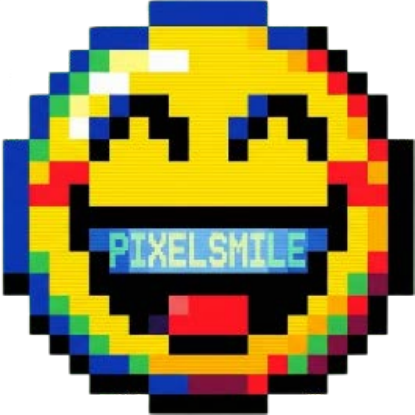}} \ourmethod{}: Toward Fine-Grained Facial Expression Editing} 

\author{
    Jiabin Hua$^{1,2,*}$
    \quad
    Hengyuan Xu$^{1,2,*}$
    \quad
    Aojie Li$^{2,\dag}$
    \quad
    Wei Cheng$^{2}$ \\
    \quad
    Gang Yu$^{2,\ddag}$ 
    \quad
    Xingjun Ma$^{1,\ddag}$
    \quad
    Yu-Gang Jiang$^{1}$\\ [0.4em]
    $^{1}$ {Fudan University}
    \quad
    $^{2}$ {StepFun} 
    \\[0.4em]
    {
        \normalsize
        \raisebox{-0.2\height}{\includegraphics[height=0.5cm]{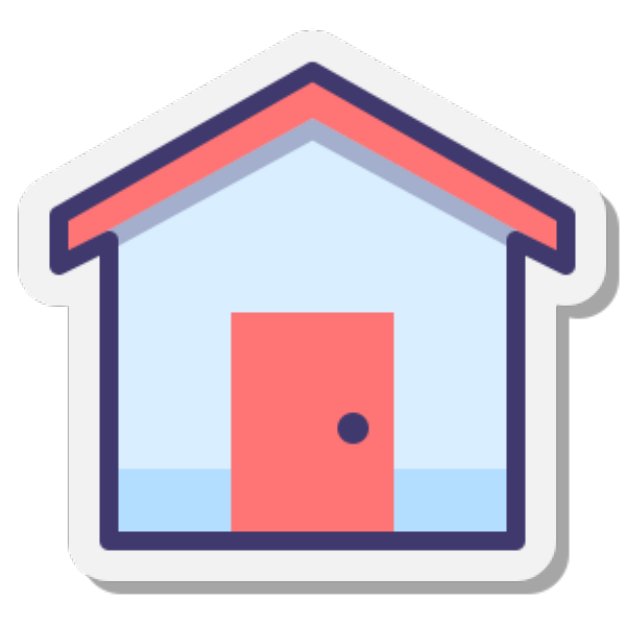}}~{\href{https://ammmob.github.io/PixelSmile/}{\textbf{Project Page}}}
        \quad
        \raisebox{-0.2\height}{\includegraphics[height=0.5cm]{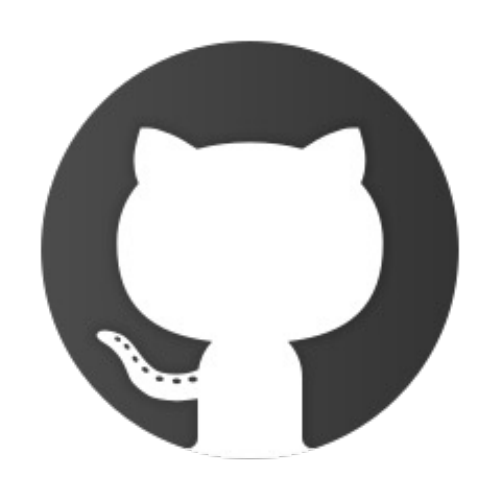}}~{\href{https://github.com/Ammmob/PixelSmile}{\textbf{Code}}}
        \quad
        \raisebox{-0.2\height}{\includegraphics[height=0.5cm]{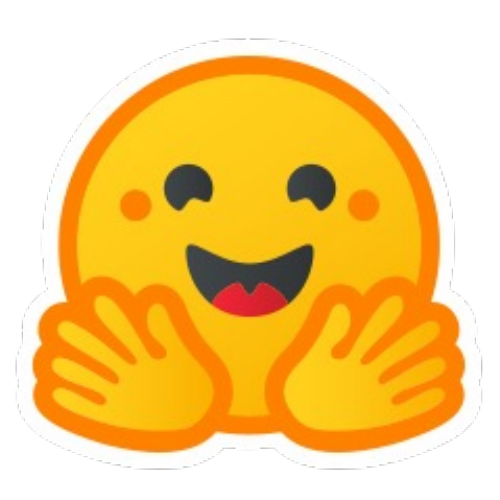}}~{\href{https://huggingface.co/PixelSmile/PixelSmile}{\textbf{Model}}}
        \quad
        \raisebox{-0.2\height}{\includegraphics[height=0.5cm]{files/logo_huggingface.pdf}}~{\href{https://huggingface.co/datasets/PixelSmile/FFE-Bench}{\textbf{Benchmark}}}
        \quad
        \raisebox{-0.2\height}{\includegraphics[height=0.5cm]{files/logo_huggingface.pdf}}~{\href{https://huggingface.co/spaces/PixelSmile/PixelSmile-Demo}{\textbf{Demo}}}
    }
}

\begin{document}
\twocolumn[{
  \renewcommand\twocolumn[1][]{#1}
  \maketitle
  \begin{center}
  \vspace{-2ex}
  \includegraphics[width=\textwidth]{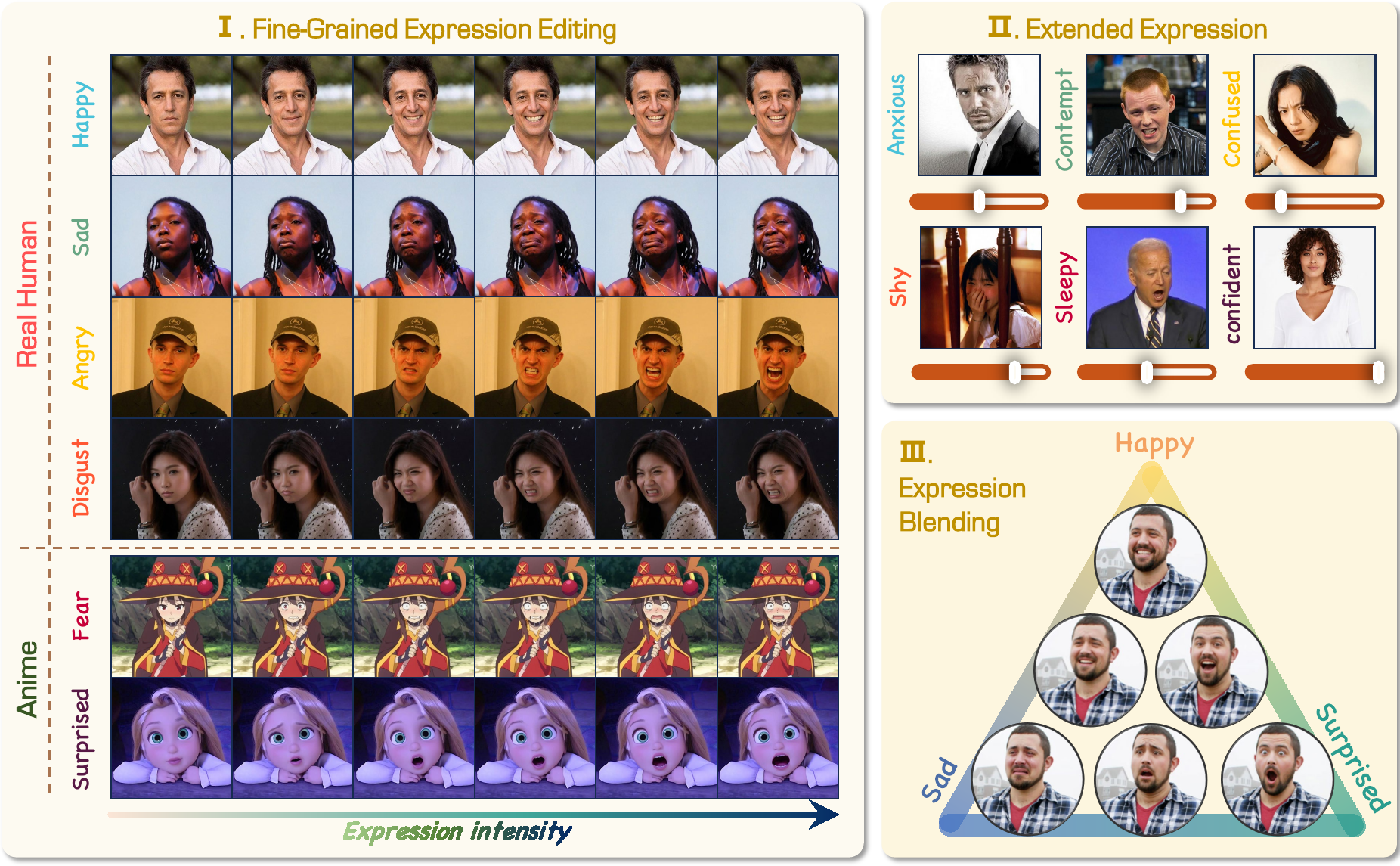}
  \captionof{figure}{\textbf{Overview of \ourmethod{}}. It enables 1) continuous and precise control of facial expression intensity across real-world and anime domains, 2) editing across 12 distinct expression categories, and 3) seamless blending of multiple expressions.}
  \label{fig:teaser}
  \end{center}
}]

{\let\thefootnote\relax\footnotetext{$^*$ Equal contribution. \dag~Project lead. \ddag~Corresponding authors.}}

\begin{abstract} Fine-grained facial expression editing has long been limited by intrinsic semantic overlap. To address this, we construct the \textbf{Flex Facial Expression} (\ourdataset{}) dataset with continuous affective annotations and establish \textbf{\ourbenchmark{}} to evaluate structural confusion, editing accuracy, linear controllability, and the trade-off between expression editing and identity preservation. We propose \textbf{\ourmethod{}}, a diffusion framework that disentangles expression semantics via fully symmetric joint training. \ourmethod{} combines intensity supervision with contrastive learning to produce stronger and more distinguishable expressions, achieving precise and stable linear expression control through textual latent interpolation. Extensive experiments demonstrate that \ourmethod{} achieves superior disentanglement and robust identity preservation, confirming its effectiveness for continuous, controllable, and fine-grained expression editing, while naturally supporting smooth expression blending.
\end{abstract}

\section{Introduction}
\label{sec:introduction}

Recent advances in diffusion-based image editing models~\cite{liu2025step1x, wu2025qwenimagetechnicalreport} and identity-consistent generation techniques~\cite{xu2025withanyone, guo2024pulid, jiang2025infiniteyou} have significantly improved the ability to manipulate personal portraits using natural language. Despite this progress, fine-grained facial expression editing remains a challenging problem. Current models can generate clearly distinct expressions, such as happy versus sad, but struggle to delineate highly correlated, semantically overlapping expression pairs, such as fear versus surprise or anger versus disgust. Most existing methods rely on discrete expression categories, forcing inherently continuous human expressions into rigid class boundaries. As a result, these formulations fail to capture subtle expression boundaries, leading to structured cross-category confusion, limited control over expression intensity, and degraded identity consistency during editing.

\begin{figure*}[t]
    \vspace{-5mm}
    \centering
    \includegraphics[width=0.9\textwidth]{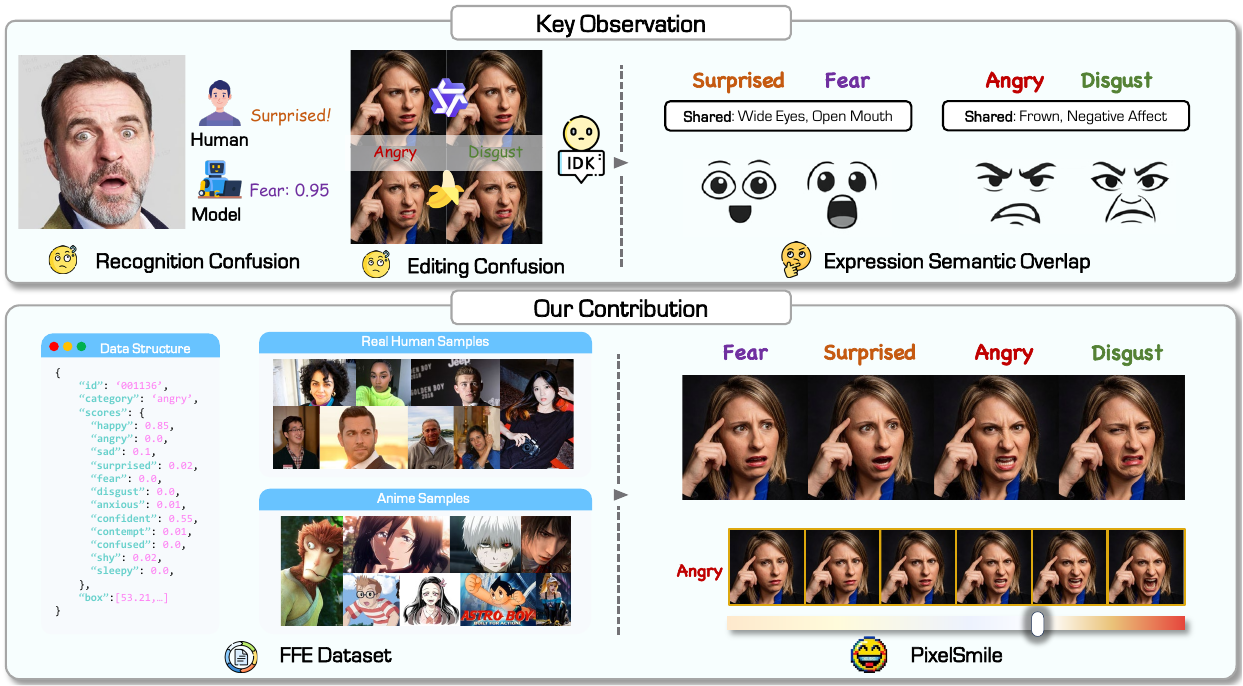}
    \caption{\textbf{Observation of Expression Semantic Overlap.} Inherent expression overlap causes systematic confusion across human annotators, recognition models, and generative models (top). We resolve this via the \ourdataset{} dataset (bottom left) and \ourmethod{} framework (bottom right), utilizing continuous supervision and symmetric training for disentangled editing.}
    \label{fig:observation}
\end{figure*}

To better understand this limitation, we analyze the semantic structure of facial expressions. As illustrated in Fig.~\ref{fig:observation}, facial expressions lie on a continuous semantic manifold where semantically adjacent emotions naturally overlap. This overlap manifests as systematic confusion across multiple stakeholders: human annotators, classifiers, and generative models often fail to uniquely distinguish semantically adjacent expressions like fear versus surprise or anger versus disgust. When generative models are trained using discrete and potentially conflicting labels from such ambiguous samples, they are forced to learn entangled representations in the latent space. Consequently, this structural entanglement prevents precise control, resulting in unintended expression leakage, where editing one emotion inadvertently triggers the characteristics of another or even degrades identity consistency.

Addressing this challenge requires a new supervision paradigm for facial expression editing models. Conventional datasets often represent facial expressions using rigid one-hot labels, which fail to capture the nuanced structure of human affect and propagate semantic entanglement into the generative pipeline. To address this limitation, we introduce a new supervision paradigm based on continuous affective annotations. Specifically, we construct the \textbf{Flex Facial Expression} (\ourdataset{}) dataset, which replaces discrete labels with continuous 12-dimensional affective score distributions. Based on this dataset, we further establish \textbf{\ourbenchmark{}} to evaluate structural confusion, editing accuracy, linear controllability, and the trade-off between expression editing and identity preservation. By providing diverse expressions within the same identity and continuous affective ground truth across both real and anime domains, \ourdataset{} breaks the one-hot supervision bottleneck, allowing models to learn the fine-grained boundaries of the expression manifold rather than disjoint categories, and enabling systematic evaluation of controllable expression editing.

Building upon this data-centric foundation, we propose \textbf{\ourmethod{}}, a diffusion-based editing framework that disentangles expression semantics. Our framework introduces a fully symmetric joint training paradigm to contrast confusing expression pairs identified in our analysis. Combined with a flow-matching-based textual latent interpolation mechanism, \ourmethod{} enables precise and linearly controllable expression intensity at inference time without requiring reference images. Through the synergy between continuous affective supervision and symmetric learning, \ourmethod{} achieves robust and controllable editing while preserving identity fidelity.

In summary, our contributions are threefold:
\begin{itemize}
    \item \textbf{Systematic Analysis of Semantic Overlap.} We reveal and formalize the structured semantic overlap between facial expressions, demonstrating that structured semantic overlap, rather than purely classification error, is a primary cause of failures in both recognition and generative editing tasks.
    \item \textbf{Dataset and Benchmark.} We construct the \ourdataset{} dataset—a large-scale, cross-domain collection featuring 12 expression categories with continuous affective annotations—and establish \ourbenchmark{}, a multi-dimensional evaluation environment specifically designed to evaluate structural confusion, expression editing accuracy, linear controllability, and the trade-off between expression editing and identity preservation.
    \item \textbf{\ourmethod{} Framework.} We propose a novel diffusion-based framework utilizing fully symmetric joint training and textual latent interpolation. This design effectively disentangles overlapping emotions and enables disentangled and linearly controllable expression editing.
\end{itemize}

\section{Related Work}
\label{sec:related_works}

\noindent\textbf{Facial Expression Editing.}
Facial expression editing aims to modify facial expressions while preserving identity. Early approaches relied on conditional GANs~\cite{goodfellow2014generative}, formulating the task as multi-domain image-to-image translation~\cite{choi2018stargan, pumarola2018ganimation, liu2019stgan, ding2018exprgan, choi2020stargan}. Subsequent works explored disentangled latent manipulation within StyleGAN-based architectures~\cite{karras2019style, karras2020analyzing, shen2020interpreting, harkonen2020ganspace, shen2021closed, yuksel2021latentclr} to identify semantic directions for continuous expression control. Another line of research incorporates explicit facial priors, such as Action Units or 3DMM parameters, to enable structured, interpretable manipulation. For instance, MagicFace~\cite{wei2025magicface} leverages such priors to guide diffusion models, while other works~\cite{pumarola2018ganimation, ding2023diffusionrig, jang2025controlface, garau2021deca, danvevcek2022emoca} explore similar structural constraints. Despite facilitating discrete expression transfers, these methods often struggle with fine-grained control, identity consistency, and generalization.
More recently, diffusion models~\cite{ho2020denoising} have significantly advanced image generation and editing quality~\cite{meng2021sdedit, hertz2022prompt, brooks2023instructpix2pix, zhang2023adding}. Furthermore, large-scale multimodal pretraining has fueled significant advancements in general-purpose editing. Large-scale foundation models, such as GPT-Image~\cite{openai_gpt_image15}, Nano Banana Pro~\cite{google_nano_banana_pro}, Qwen-Image~\cite{wu2025qwenimagetechnicalreport}, and LongCat-Image~\cite{LongCat-Image}, now demonstrate remarkable zero-shot flexibility and editing capabilities~\cite{bytedance_seedream45, flux-2-2025, liu2025step1x}.

\noindent\textbf{Continuously Controlled Generation.}
Prior works achieve continuous editing by leveraging interpolatable subspaces within generative models. 
ConceptSlider~\cite{gandikota2024concept} interpolates LoRA weights, while subsequent methods~\cite{baumann2025continuous, guerrero2024texsliders, sridhar2024prompt, garibi2025tokenverse, zhong2025mod, kamenetsky2025saedit, jain2025adaptivesliders, ye2025all, dalva2024fluxspace,gandikota2025sliderspace,hertz2022prompt,sharma2024alchemist,cheng2025marble} manipulate text embeddings or modulation features to achieve gradual semantic variation. More recently, SliderEdit~\cite{zarei2025slideredit}, Kontinuous-Kontext~\cite{parihar2025kontinuous}, and concurrent works~\cite{wolf2026continuous, xu2025numerikontrol, yin2025instructattribute} extend continuous control to editing models built upon FLUX.1 Kontext~\cite{batifol2025flux}. 
Despite smoother transitions via reduced strength or pixel interpolation, these methods remain constrained by entangled latent spaces, leading to semantic ambiguity and identity drift at large magnitudes.
By disentangling latent expression semantics, our structured formulation achieves fine-grained linear control and identity preservation across diverse manipulation strengths.

\noindent\textbf{Facial Expression Datasets and Benchmarks.}
High-quality datasets and reliable benchmarks are essential for facial expression analysis. 
Early controlled datasets~\cite{langner2010presentation, lundqvist1998karolinska, lucey2010extended, yin20063d} provide same-identity multi-expression samples for precise comparison but lack diversity, while large-scale in-the-wild datasets~\cite{mollahosseini2017affectnet, li2017reliable, BarsoumICMI2016, zhang2018facial,yang2025controllable} enhance generalization but lack paired expressions for the same identity, hindering identity-expression disentanglement in generative editing.
Recent efforts extend to video and multimodal settings. While video-based datasets~\cite{nagrani2020voxceleb, qiu2025emovid, zhang2025videmo} focus on temporal or cross-modal dynamics, the MEAD dataset~\cite{wang2020mead} provides expressions with three distinct intensity levels, moving beyond purely categorical labels but still falling short of fine-grained, continuous control and structured disentanglement in static editing contexts. Alongside these, benchmarks such as F-Bench~\cite{liu2025f} and SEED~\cite{zhu2025seed} evaluate facial generation using visual metrics and human preference. However, standard metrics (e.g., CLIP, SSIM, LPIPS) capture overall quality but offer limited insight into disentanglement and continuous control.
To address these gaps, we propose \ourdataset{} and \ourbenchmark{}. By providing same-identity pairs with continuous affective annotations, our approach enables rigorous evaluation of fine-grained, linearly controllable, and disentangled expression editing.
\begin{figure*}[htbp]
    \vspace{-5mm}
    \centering
    \includegraphics[width=\textwidth]{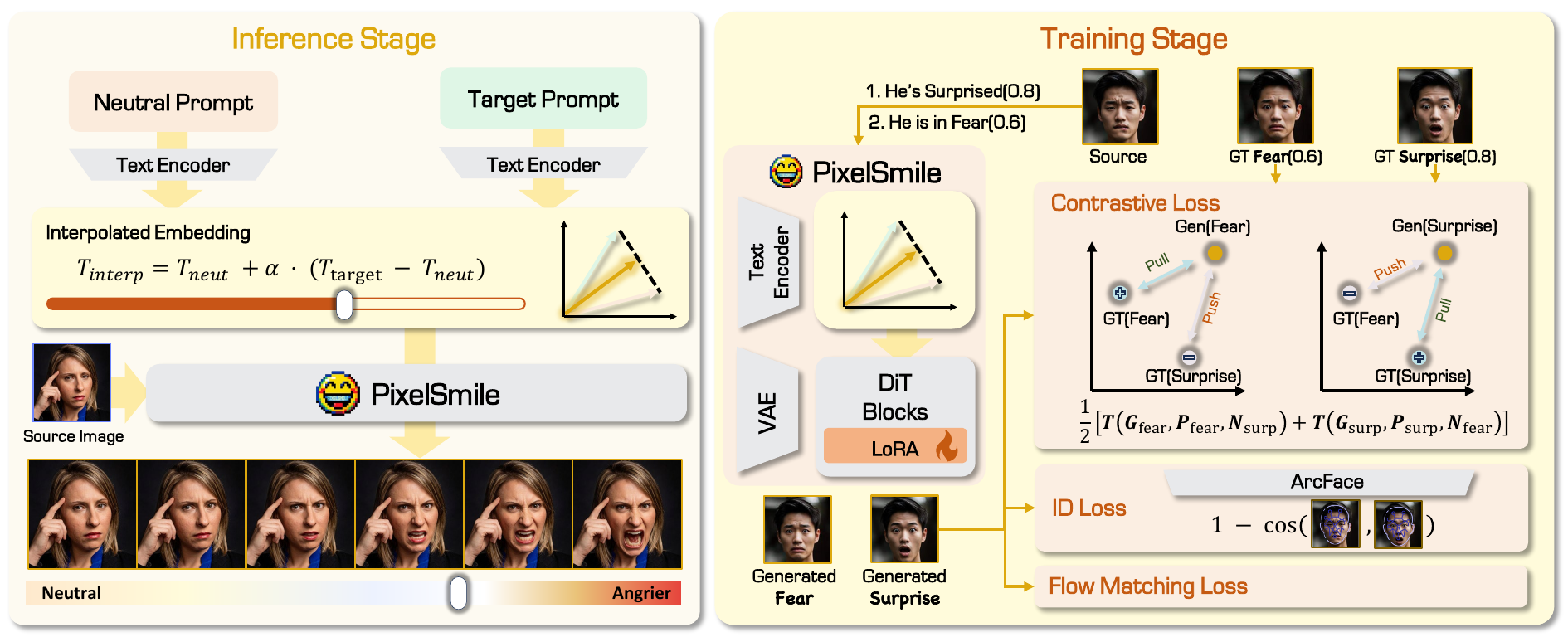}
\caption{\textbf{Framework Overview.} (1) \textbf{Inference Stage}. We interpolate between the neutral and target expression embeddings in textual latent space using a controllable coefficient $\alpha$, enabling continuous adjustment of expression intensity. (2) \textbf{Training Stage}. We adopt a joint fully symmetric training framework. Specifically, we sample a source image $P_{\mathrm{src}}$ and a confusing expression pair $(P_a, P_b)$ to construct a triplet. We first treat $P_a$ as the positive and $P_b$ as the negative to compute a joint loss, and then swap their roles to compute it again, yielding a symmetric training objective. The joint loss consists of three components: a Flow-Matching loss for intensity alignment, a contrastive loss for expression separation, and an identity preservation loss to maintain subject consistency.}
    \label{fig:framework}
\end{figure*}
\section{Dataset and Benchmark}
\label{sec:dataset}

To facilitate fine-grained and linearly controllable facial expression editing, we construct the \ourdataset{} dataset and establish \ourbenchmark{}, a dedicated evaluation benchmark. 
Existing datasets often lack same-identity expression diversity or provide only discrete expression labels, which limits the evaluation of controllable expression manipulation. Our dataset addresses these limitations by providing large-scale same-identity expression variations with continuous affective annotations, enabling systematic analysis of expression disentanglement and editing controllability.

\subsection{The \ourdataset{} Dataset}

\ourdataset{} is constructed through a four-stage \emph{collect--compose--generate--annotate} pipeline designed to ensure expression diversity, cross-domain coverage, and reliable annotations. 
The final dataset contains 60,000 images across real and anime domains, supporting both photorealistic and stylized facial expression editing.

\noindent\textbf{Base Identity Collection.}
We first curate a set of high-quality base identities from two domains: 
(1) \emph{Real domain}: approximately 6,000 real-world portraits are collected from public portrait datasets~\cite{snmahsa_human_images_dataset, matting_human_datasets}, covering diverse demographics and scene compositions, including both close-up and full-body images; 
(2) \emph{Anime domain}: to enable cross-domain evaluation, we collect stylized portraits from 207 anime productions covering 629 characters, from which around 6,000 high-quality images are retained after quality filtering and automated face detection. 
For both domains, automated face detection followed by manual verification is applied to ensure identity clarity and image quality. 
These images form the identity backbone of \ourdataset{} dataset.

\noindent\textbf{Expression Prompt Composition.}
To obtain fine-grained expression variations, we construct a structured prompt library for 12 target expressions. 
The taxonomy consists of six basic emotions~\cite{ekman1992argument} and six extended emotions (Confused, Contempt, Confident, Shy, Sleepy, Anxious). Rather than relying solely on abstract expression labels, each expression is decomposed into facial attribute components (e.g., mouth shape, eyebrow movement, and eye openness).
Candidate attribute combinations are automatically generated and filtered with a vision-language model to remove anatomically inconsistent or semantically conflicting descriptions, resulting in a validated library of fine-grained expression prompts.

\noindent\textbf{Controlled Expression Generation.}
For each base identity, multiple target expressions with varying intensities are synthesized using a state-of-the-art image editing model, \emph{\nbp{}}. 
We adopt a dual-part prompt design that specifies both the global expression category and localized facial attributes, improving controllability and reducing ambiguity between semantically similar expressions.
This process produces approximately 60,000 images in total (30,000 per domain), providing rich identity-preserving expression variations across diverse conditions.

\noindent\textbf{Continuous Annotation and Quality Filtering.}
Departing from conventional one-hot expression labels, each image is annotated with a 12-dimensional continuous score vector $\mathbf{v} \in [0,1]^{12}$. 
The scores are predicted by a vision-language model, \emph{Gemini 3 Pro}, which estimates the intensity of each expression category. A subset of samples is verified by human annotators to ensure reliability.
This representation captures semantic overlap between facial expressions (e.g., fear and surprise), providing a faithful approximation of the affective manifold. 
We further perform consistency checks and manual spot verification to remove ambiguous or low-confidence samples. 
The resulting dataset provides same-identity expression variations with continuous soft labels, enabling fine-grained evaluation of expression disentanglement and controllable facial expression editing.

\subsection{The \ourbenchmark{} Benchmark}
\label{sec:dataset_benchmark}

Motivated by the intrinsic semantic entanglement among facial expressions, 
which leads to structured cross-category confusion, 
we design a unified benchmark to evaluate facial expression editing from four complementary aspects: 
structural confusion, the trade-off between expression editing and identity preservation, control linearity, and expression editing accuracy.
All expression classifications and intensity scores are predicted by Gemini 3 Pro.

\noindent\textbf{Mean Structural Confusion Rate (mSCR).}
To quantify structured confusion between semantically similar expressions, we define the directed confusion rate $C_{i \rightarrow j}$ and the bidirectional confusion rate (BCR) as follows:
\begin{align}
C_{i \rightarrow j} &= \frac{1}{N_i} \sum_{k=1}^{N_i} \mathbf{1}(\hat{y}_k^{(i)} = j), \\
\mathrm{BCR}(i,j) &= \frac{1}{2}\left(C_{i \rightarrow j} + C_{j \rightarrow i}\right),
\end{align}
where $N_i$ denotes the number of samples edited toward class $i$, and $\hat{y}_k^{(i)}$ is the predicted dominant expression. The mSCR is computed by averaging $\mathrm{BCR}(i,j)$ over predefined confusing pairs (e.g., Fear--Surprise and Angry--Disgust). A lower mSCR indicates reduced cross-category confusion and improved semantic disentanglement.

\noindent\textbf{Harmonic Editing Score (HES).}
Facial expression editing requires both accurate expression transfer and identity preservation.
We define the Harmonic Editing Score as
\begin{equation}
\mathrm{HES} = 
\frac{2 \times S_E \times S_{\mathrm{ID}}}{S_E + S_{\mathrm{ID}}},
\end{equation}
where \(S_E\) denotes the VLM-based target expression score, 
and \(S_{\mathrm{ID}}\) is the cosine similarity between source and edited faces. 
Identity similarity is computed as the average cosine similarity from three face recognition models
(including ArcFace~\cite{deng2019arcface}, AdaFace~\cite{kim2022adaface}, FaceNet~\cite{schroff2015facenet}) for robustness. 
High HES is achieved only when both expression strength and identity fidelity are preserved.

\noindent\textbf{Control Linearity Score (CLS).} To evaluate continuous controllability, 
we feed uniformly spaced intensity coefficients \(\alpha \in [0,\alpha_{\max}]\) 
during inference and compute the Pearson correlation between \(\alpha\) 
and the VLM-predicted intensity scores. 
Higher CLS indicates more linear and predictable expression control.

\noindent\textbf{Expression Editing Accuracy (Acc).} We report the proportion of generated images whose predicted dominant expression matches the target instruction.
This metric measures overall categorical editing success.
\begin{figure}[htbp]
    \centering
    \includegraphics[width=0.95\linewidth]{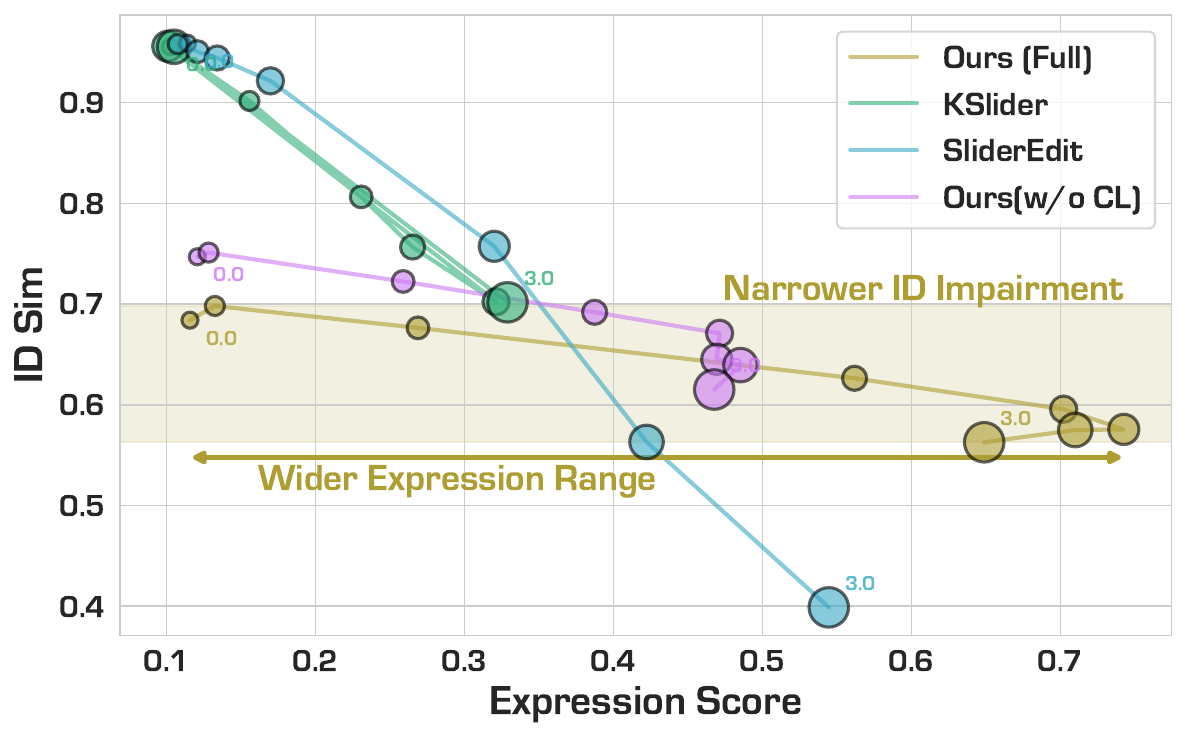}
    \caption{\textbf{Quantitative Evaluation of Linear Control Methods}. Comparison of the trade-off between ID similarity and expression score across different models. \ourmethod{} achieves an optimal balance, providing a wider expression manipulation range while preserving identity fidelity.}
    \label{fig:quantitative_linechart}
\end{figure}

\begin{table*}[hbtp]
\vspace{-5mm}
\centering
\begin{minipage}[t]{0.48\textwidth}
    \centering
    \caption{\textbf{Quantitative Evaluation of General Editing Models}. Best, second best, and third best results are indicated by \legendsquare{colorbest}, \legendsquare{colorsecond}, and \legendsquare{colorthird} respectively.}
    \label{tab:quantitative_table1}
    \resizebox{\linewidth}{!}{
    \begin{tabular}{l|l|cccc}
    \toprule
    & Method & mSCR $\downarrow$ & Acc-6 $\uparrow$ & Acc-12 $\uparrow$ & ID Sim $\uparrow$ \\ 
    \midrule
    \multirow{3}{*}{Closed} 
    & \seed~\cite{bytedance_seedream45} & 0.3725 & 0.5294 & 0.3737 & \best 0.7221 \\
    & Nano Banana Pro~\cite{google_nano_banana_pro} & 0.1754 & \second 0.8431 & \second 0.6200 & \second 0.7107 \\
    & \gpt~\cite{openai_gpt_image15} & \second 0.1107 & \third 0.8039 & \best 0.6300 & 0.5056 \\ 
    \midrule
    \multirow{4}{*}{Open} 
    & \flux~\cite{flux-2-2025} & 0.2850 & 0.4510 & 0.3310 & 0.4146 \\
    & \longcat~\cite{LongCat-Image} & \third 0.1754 & 0.6275 & 0.4100 & \third 0.6036 \\
    &\qwen~\cite{wu2025qwenimagetechnicalreport} & 0.2625 & 0.4510 & 0.2900 &0.6938 \\
    & Ours(w/o training) & 0.2400 & 0.5294 & 0.3500 & 0.6769 \\ 
    & Ours & \best 0.0550 & \best 0.8627 & \third 0.6000 & 0.6522 \\ 
    \bottomrule
    \end{tabular}
    }
\end{minipage}
\hfill
\begin{minipage}[t]{0.48\textwidth}
    \centering
    \caption{\textbf{Quantitative Evaluation of Linear Control Models}. Best, second best, and third best results are indicated by \legendsquare{colorbest}, \legendsquare{colorsecond}, and \legendsquare{colorthird} respectively.}
    \label{tab:quantitative_table2}
    \resizebox{\linewidth}{!}{
    \begin{tabular}{l|cccc}
    \toprule
    Method & CLS-6 $\uparrow$ & CLS-12 $\uparrow$ & ID Sim $\uparrow$ & \textbf{HES}$\uparrow$ \\ 
    \midrule
    SAEdit~\cite{kamenetsky2025saedit} & -0.0183 & 0.0007 & - & - \\
    ConceptSlider~\cite{gandikota2024concept} & 0.3161$^{\ast}$ & - & \third 0.6250 & \third 0.3656 \\
    AttributeControl~\cite{baumann2025continuous} & 0.2856$^{\ast}$ & - & 0.3609 & 0.2712 \\
    \kslider~\cite{parihar2025kontinuous} & -0.0459 & -0.0634 & \best 0.7974 & 0.3272 \\
    SliderEdit~\cite{zarei2025slideredit} & \third 0.5599 & \second 0.5217 & \second 0.7414 & 0.3441 \\
    Ours(w/o training) & \second 0.6892 & \second 0.5217 & 0.6769 & \second 0.4086 \\ 
    Ours & \best 0.8078 & \best 0.7305 & 0.6522 & \best 0.4723 \\ 
    \bottomrule
    \multicolumn{5}{l}{\footnotesize $^{\ast}$ Evaluated on CLS-2 (\textit{happy, surprised}).} \\
    \end{tabular}
    }
\end{minipage}
\vspace{-5mm}
\end{table*}
\section{Method}
\label{sec:method}

We present \ourmethod{}, a framework for fine-grained facial expression editing. 
As illustrated in Fig.~\ref{fig:framework}, our method builds upon a pretrained 
Multi-Modal Diffusion Transformer (MMDiT)~\cite{peebles2023scalable} 
with LoRA adaptation~\cite{hu2022lora}.
To address intrinsic semantic entanglement and enable continuous intensity control, 
we introduce two key components: 
(1) a Flow-Matching-based textual interpolation mechanism~\cite{lipman2022flow} 
for smooth expression strength control; and 
(2) a Fully Symmetric Joint Training framework with a symmetric contrastive objective 
to reduce cross-category confusion while preserving identity and background consistency.

\subsection{Textual Latent Interpolation for Continuous Editing}

Existing expression editing approaches typically rely on discrete labels 
or coarse reference signals~\cite{wu2025qwenimagetechnicalreport}, 
which limits fine-grained control over expression intensity. 
Instead, we perform linear interpolation in the textual latent space 
to enable continuous and smooth expression manipulation.

\noindent\textbf{Textual Latent Interpolation.} Given a neutral prompt \(P_{\mathrm{neu}}\) and a target expression prompt \(P_{\mathrm{tgt}}\), 
the frozen MMDiT text encoder maps them to embeddings 
\(e_{\mathrm{neu}}\) and \(e_{\mathrm{tgt}}\), respectively. 
We define the residual direction
\begin{equation}
\Delta e = e_{\mathrm{tgt}} - e_{\mathrm{neu}},
\end{equation}
which captures the semantic shift from neutral to the target expression.

A continuous conditioning embedding is then constructed as
\begin{equation}
e_{\mathrm{cond}}(\alpha) = 
e_{\mathrm{neu}} + \alpha \cdot \Delta e,
\quad \alpha \in [0,1].
\end{equation}
When \(\alpha=0\), the conditioning corresponds to neutral expression; when \(\alpha=1\), it recovers the full target expression. Intermediate values of \(\alpha\) yield smoothly varying expression intensities. Importantly, the same direction also supports extrapolation: at inference time, \(\alpha > 1\) enables stronger expression transfer while maintaining structural consistency.

\noindent\textbf{Score-Supervised Flow Matching.} To enforce consistency between textual interpolation and visual intensity, we introduce score supervision during Flow Matching (FM) training. Each training image is associated with a ground-truth intensity coefficient \(\alpha_{\mathrm{gt}} \in [0,1]\), derived from the continuous expression annotations. During LoRA fine-tuning, we set \(\alpha=\alpha_{\mathrm{gt}}\) and use \(e_{\mathrm{cond}}(\alpha)\) as the conditioning input to the dual-stream attention blocks. The score-supervised velocity loss is defined as
\begin{equation}
\mathcal{L}_{\mathrm{FM}}^{\mathrm{edit}} =
\mathbb{E}_{t,x_0,x_1}
\Big[
\big\|
v_\theta(x_t, t, e_{\mathrm{cond}}(\alpha))
- (x_1 - x_0)
\big\|_2^2
\Big],
\end{equation}
where $x_0$ denotes the source image latent and $x_1$ denotes the edited target latent.
This objective explicitly couples the interpolation coefficient 
with the corresponding visual transformation. 
At inference, continuous control is achieved by varying \(\alpha\), 
without requiring reference images.

\subsection{Fully Symmetric Joint Training for Disentanglement}

As stated in Sec.~\ref{sec:introduction} and illustrated in Fig.~\ref{fig:observation}, 
facial expressions lie on a continuous and highly overlapping semantic manifold. 
For example, \textit{Surprise} and \textit{Fear} share similar arousal and facial cues, 
leading to structural confusion near class boundaries when trained with discrete supervision only. 
Inspired by contrastive learning and the idea of symmetric learning~\cite{wang2019symmetric}, 
we introduce a Fully Symmetric Joint Training framework with a symmetric contrastive objective in the feature space.

\noindent\textbf{Symmetric Construction.} 
Given a pair of semantically overlapping expressions, $(E_a, E_b)$, defined based on the confusion patterns observed in the FFE dataset, and an input image, the model performs two parallel generations, $G_a$ and $G_b$, conditioned on prompts corresponding to $E_a$ and $E_b$, respectively. 
For $G_a$, the ground-truth image with expression $E_a$, denoted as $P_a$, serves as the positive, while the image with expression $E_b$, denoted as $P_b$, is treated as a hard negative; the roles are reversed for $G_b$. 
This symmetric design avoids directional bias and enforces consistent separation between confusing expressions.

\noindent\textbf{Symmetric Contrastive Loss.} All images are encoded using a frozen CLIP image encoder to capture expression semantics.
The symmetric loss is defined as
\begin{equation}
\mathcal{L}_{\mathrm{SC}} =
\frac{1}{2}
\left[
\mathcal{T}(G_a, P_a, P_b)
+
\mathcal{T}(G_b, P_b, P_a)
\right],
\end{equation}
where $\mathcal{T}$ pulls the generated sample toward its target while pushing it away from the confusing expression.

We investigate three realizations of $\mathcal{T}$, including hinge-based \cite{schroff2015facenet}, log-ratio \cite{oh2016deep}, and InfoNCE-style \cite{oord2018representation} formulations. In practice, we primarily adopt the InfoNCE-style objective due to its stable optimization. Detailed formulations and ablations are provided in the Appendix \ref{app:appendix_sc}.

\subsection{Identity Preservation}

Strong intensity extrapolation ($\alpha>1$) or contrastive forces may degrade identity consistency. To stabilize biometric features, we introduce an identity preservation loss based on a pretrained face recognition model. Specifically, we adopt ArcFace \cite{deng2019arcface} as a frozen identity encoder $\Phi_{\mathrm{arc}}(\cdot)$. For generated images $G_a, G_b$ and their corresponding 
ground truths $P_a, P_b$, the identity loss is defined as

\begin{equation}
\mathcal{L}_{\mathrm{ID}} =
\frac{1}{2}\sum_{i \in \{a,b\}} \left[ 1 - \cos(\Phi_{\mathrm{arc}}(G_i), \Phi_{\mathrm{arc}}(P_i)) \right],
\end{equation}

This term enforces identity consistency while allowing expression variation.

\subsection{Overall Training Objective}

We fine-tune the LoRA parameters of the frozen MMDiT under a symmetric dual-branch training scheme, where a pair of confusing expressions $(a,b)$ is optimized jointly for the same subject. The overall objective is defined as
\begin{equation}
\mathcal{L}_{\mathrm{total}} =
\frac{1}{2}\!\left(
\mathcal{L}_{\mathrm{FM}}^{a}
+
\mathcal{L}_{\mathrm{FM}}^{b}
\right)
+ \lambda_{\mathrm{sc}} \mathcal{L}_{\mathrm{SC}}
+ \lambda_{\mathrm{id}} \mathcal{L}_{\mathrm{ID}},
\end{equation}
where $\lambda_{\mathrm{sc}}$ and $\lambda_{\mathrm{id}}$ control the trade-off between disentanglement and identity preservation. This symmetric formulation jointly enforces 
continuous intensity control, expression separation, 
and identity consistency.
\begin{figure*}[htbp]
    \vspace{-5mm}
    \centering
    \includegraphics[width=0.9\textwidth]{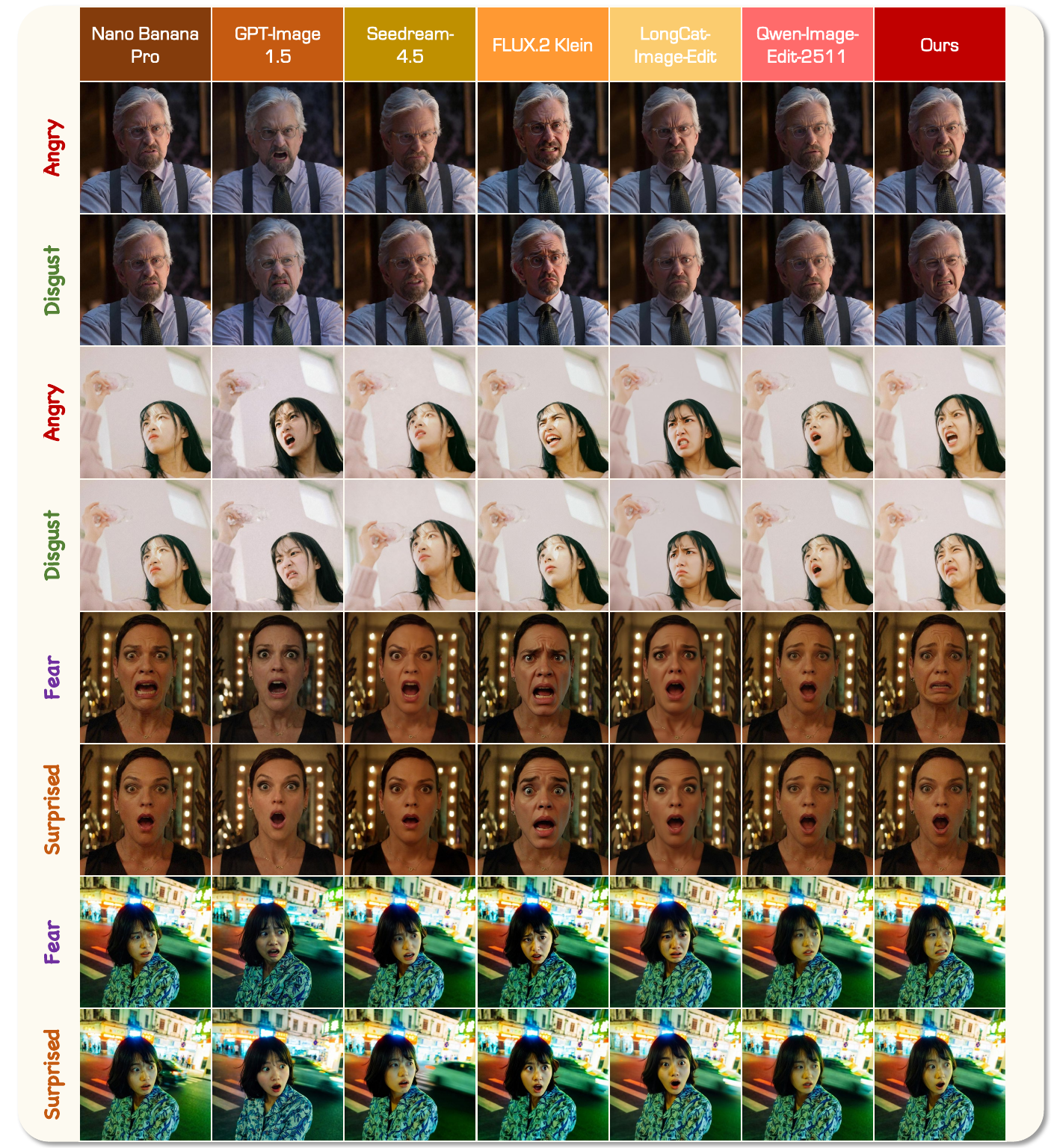}
    \caption{\textbf{Qualitative Comparison with General Editing Models.} 
    \ourmethod{} produces clearer expression changes while preserving facial identity, 
    whereas existing editing models either weaken expression editing or degrade identity consistency.}
    \label{fig:quality_general}
\end{figure*}
 
\begin{figure*}[htbp]
    \vspace{-5mm}
    \centering
    \includegraphics[width=0.9\textwidth]{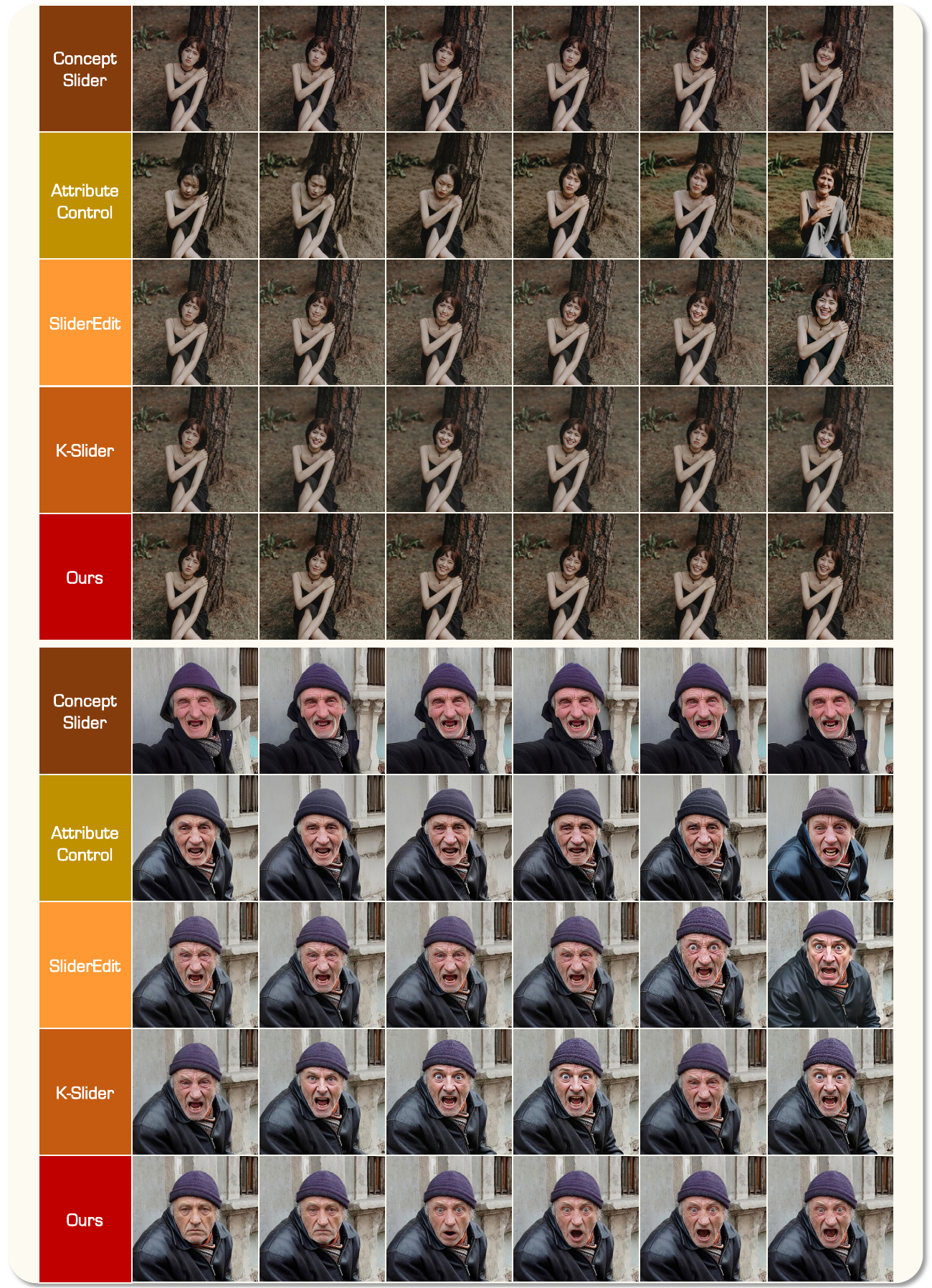}
    \caption{\textbf{Qualitative Comparison with Linear Control Models.} \ourmethod{} achieves smooth and monotonic expression transitions while preserving facial identity, whereas existing control methods either produce unstable responses or sacrifice identity consistency. The figure illustrates two representative expressions: happy (top row) and surprised (bottom row).}
    \label{fig:quality_linear}
\end{figure*}

\begin{figure}[hbp]
    \centering
    \includegraphics[width=0.8\linewidth]{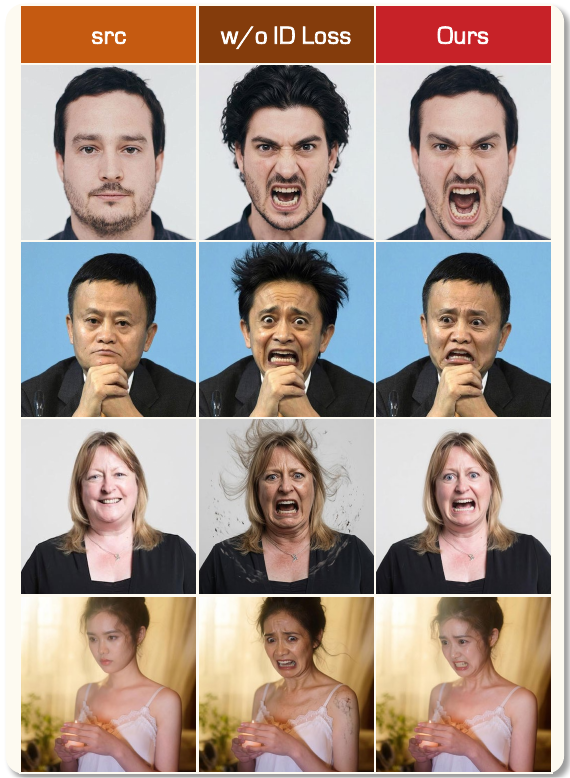}
    \caption{\textbf{Ablation on identity loss.} Without ID loss, large expression intensities cause identity drift in hairstyle and skin texture. Our full method preserves identity consistently.}
    \label{fig:ablation_idloss}
    \vspace{-5mm}
\end{figure}
\begin{figure*}[htbp]
    \centering
    \includegraphics[width=0.9\textwidth]{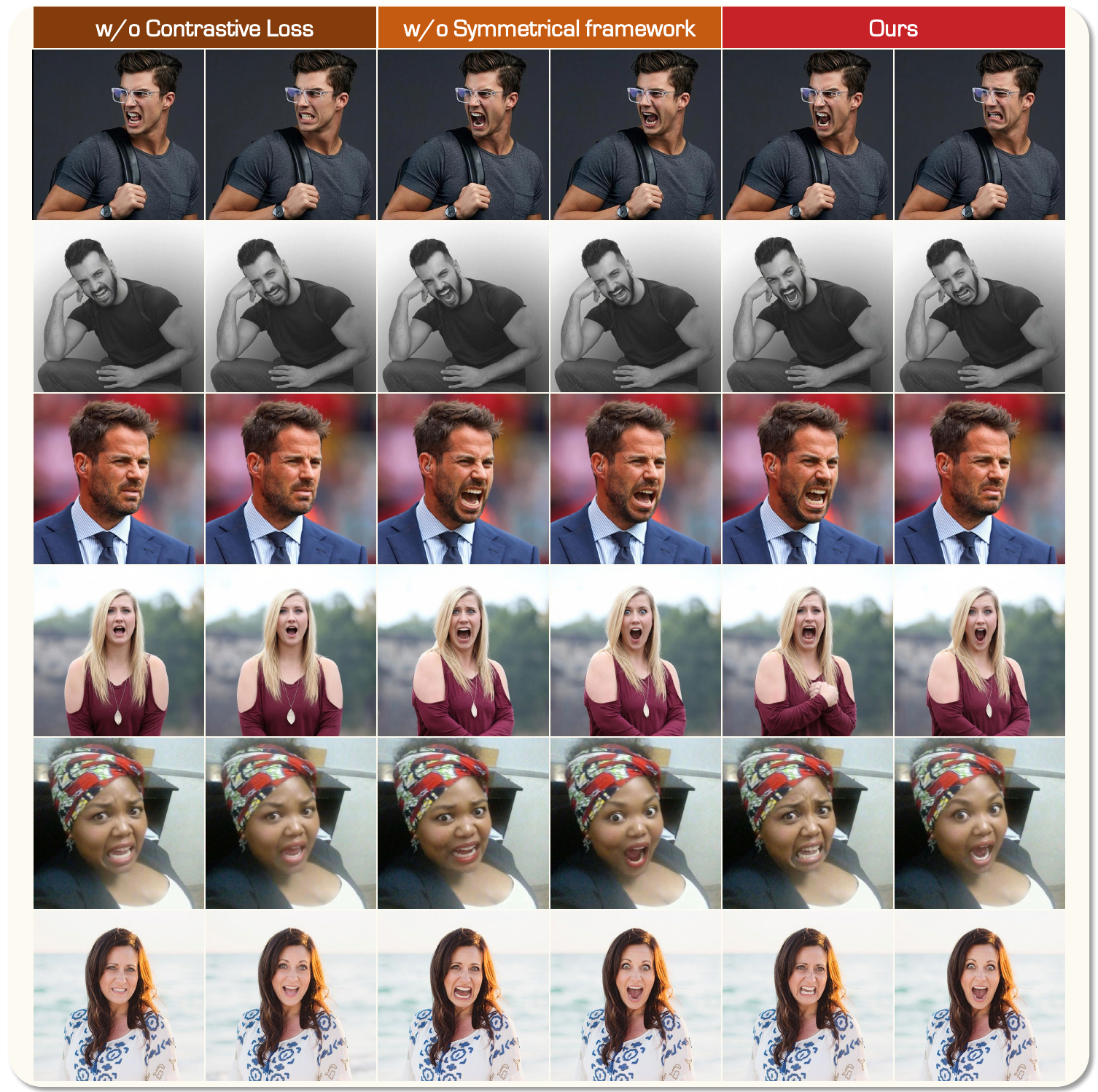}
    \caption{\textbf{Ablation on symmetric contrastive learning.} 
    Both w/o Contrastive Loss and w/o Symmetric Framework suffer from expression confusion, while our full method achieves precise expression disentanglement. 
    The upper three rows show angry and disgust, and the lower three rows show fear and surprised.}
    \label{fig:ablation_scloss}
\end{figure*}

\section{Experiment}
\label{sec:experiment}
\subsection{Experimental Setup}

We implement \ourmethod{} based on Qwen-Image-Edit-2511. 
To handle the distinct stylistic distributions of real-world and anime domains, 
we train two independent LoRA adapters for each. Following prior work~\cite{zhuang2025vistorybench,chang2025oneig}, for contrastive supervision, we adopt CLIP-ViT-L/14~\cite{radford2021learning} for the real domain and DanbooruCLIP~\cite{danbooruclip} for anime. 
Identity preservation is enforced using a pretrained ArcFace (antelopev2) model 
for the real domain. Additional implementation details are provided in Appendix~\ref{app:appendix_exp}.

\noindent\textbf{Baselines.}
To ensure a comprehensive and fair evaluation, 
we divide baselines into two groups according to their primary strengths in facial expression editing: general editing models, which are strong in overall expression editing quality, and linear control models, which are designed for continuous and predictable intensity control.

\textbf{Group 1: General Editing Models.}
This group represents the strongest general-purpose text-guided image editing systems. 
We include three closed-source commercial systems: Nano Banana Pro, GPT-Image-1.5 (\gpt), Seedream-4.5 (\seed), 
and three open-source models: Qwen-Image-Edit-2511 (\qwen), FLUX.2 Klein (\flux), and LongCat-Image-Edit (\longcat). 
In the following, we refer to each model by the abbreviated name in parentheses. Although these models do not provide explicit mechanisms for fine-grained linear control, 
their strong generative priors make them competitive in overall expression editing quality. 
We therefore use them to evaluate expression editing accuracy and the ability to resolve structural confusion between semantically overlapping expressions.

\textbf{Group 2: Linear Control Models.}
This group focuses on continuous attribute manipulation in latent space. 
We compare with recent control-oriented editing models including Kontinuous-Kontext (\kslider), SliderEdit, and SAEdit. 
We also include earlier latent control approaches ConceptSlider and AttributeControl, 
using their officially recommended inversion strategies for real-image editing. 
While these earlier methods pioneered latent attribute control, they are often limited by narrow predefined attribute categories and information loss introduced by inversion. 
We therefore treat them as reference baselines rather than primary competitors in multi-category quantitative evaluation.

\noindent\textbf{Evaluation Metrics.}
We adopt the benchmark protocol defined in Sec.~\ref{sec:dataset_benchmark}. 
For Group 1, we evaluate editing accuracy and expression disentanglement using Acc-6, Acc-12, and mSCR. 
For Group 2, we evaluate linear intensity control and identity fidelity using CLS-6, CLS-12, and HES.

\subsection{Quantitative Evaluation}
We quantitatively compare \ourmethod{} with both general editing and linear control models in Table~\ref{tab:quantitative_table1} and Table~\ref{tab:quantitative_table2}.

\noindent\textbf{Evaluation with General Editing Models.}
As shown in Table~\ref{tab:quantitative_table1}, we evaluate baselines on editing accuracy, structural confusion, and identity fidelity. 
For the six basic expressions, \ourmethod{} achieves the highest editing accuracy (0.8627), surpassing \nbp\ (0.8431) and \gpt\ (0.8039). 
On the twelve extended expressions, our method remains among the best-performing models. This partially reflects the bias of the VLM scoring model (Gemini 3 Pro), which is highly reliable on basic expressions but less consistent on extended expression categories.
More importantly, \ourmethod{} achieves the lowest structural confusion rate (0.0550), significantly outperforming \gpt\ (0.1107) and \nbp\ (0.1754), while most other models exceed 0.2000. A value approaching 0.5 indicates that the model tends to collapse the confusing expression pair into a single expression, reflecting poor disentanglement of overlapping expressions.
In terms of identity fidelity, empirical observations in \cite{xu2025withanyone} suggest that realistic facial expression editing typically yields ID similarity values around 0.6–0.7. 
Scores above 0.8 often indicate rigid ``copy-paste'' behavior, while scores below 0.5 imply severe identity distortion. 
Some baselines fall into these extremes: \seed\ maintains high ID similarity but suffers from large structural confusion due to limited edits, whereas \flux\ drops below 0.5, significantly degrading identity consistency. 
In contrast, \ourmethod{} produces strong expressions while maintaining identity similarity within the natural range, achieving a better balance between expression strength and identity preservation.

\noindent\textbf{Evaluation with Linear Control Models.}
\ourmethod{} demonstrates robust and consistent linear controllability across all metrics. ConceptSlider and AttributeControl are limited to editing only two expression attributes (happy and surprised) and produce weak editing effects; therefore, we report them as reference baselines with partial metrics (e.g., CLS-2) rather than full-category comparisons. SAEdit is a text-to-image method that does not explicitly support identity-preserving editing; therefore, we include it only for quantitative reference and do not provide detailed qualitative analysis.
As shown in Table~\ref{tab:quantitative_table2} and Figure~\ref{fig:quantitative_linechart}, simply applying textual embedding interpolation to \qwen\ (zero-shot) already yields competitive controllability (CLS-6 0.6892, HES 0.4086), outperforming existing control baselines. 
With the proposed symmetric joint training, \ourmethod{} further improves performance and achieves the best results across all benchmarks (CLS-6 0.8078, CLS-12 0.7305, and HES 0.4723), indicating that explicitly modeling expression semantics is critical for stable and fine-grained controllability.
Figure~\ref{fig:quantitative_linechart} further reveals the limitations of existing methods. 
Although \kslider\ and SliderEdit maintain high average ID similarity, 
this is largely because low editing intensities produce negligible changes, yielding ID similarity values close to 1.0. Specifically, \kslider\ exhibits negative CLS scores and irregular intensity fluctuations that never exceed $\sim$0.3, failing to establish linear controllability. 
SliderEdit shows increasing expression intensity but forces a rapid drop in ID similarity (down to $\sim$0.4) once expression scores approach 0.5, indicating a trade-off between editing strength and identity preservation. 
In contrast, \ourmethod{} achieves a monotonic response across a wide intensity range (expression scores reaching $\sim$0.8) while maintaining identity similarity within the natural 0.6--0.7 interval, effectively balancing controllability and fidelity. This behavior demonstrates that our method not only improves average performance but also ensures stable and predictable control across the intensity spectrum.

\begin{table*}[htbp]
    \centering
    \caption{\textbf{Ablation Study}. Best, second best, and third best results are indicated by \legendsquare{colorbest}, \legendsquare{colorsecond}, and \legendsquare{colorthird} respectively.}
    \scalebox{0.9}{
    \begin{tabular}{l|l|ccccccc}
            \toprule
             & \textbf{Ablation} & mSCR $\downarrow$ & ACC-6 $\uparrow$ & ACC-12 $\uparrow$ & CLS-6 $\uparrow$ & CLS-12 $\uparrow$ & HES $\uparrow$ & ID Sim $\uparrow$ \\
            \midrule
            \multirow{3}{*}{Loss} 
                & w/o Contrastive Loss      & 0.2725 & 0.6471 & 0.5889 & 0.6978 & 0.5889 & 0.4500 & \best 0.7018 \\
                & w/o ID Loss               & \second 0.0550 & \second 0.8824 & \second 0.6500 & \best 0.8215 & \third 0.6874 & 0.4451 & 0.5749 \\
                & w/o Sym. Frame.           & 0.1350 & 0.7843 & 0.4700 & 0.7939 & 0.6488 & 0.4253 & 0.6402 \\
            \midrule
            \multirow{2}{*}{Constraint} 
                & w/ Log-Ratio Constraint   & 0.1750 & 0.8039 & 0.5300 & 0.7917 & 0.6546 & \best 0.4933 & \second 0.6943 \\
                & w/ Hinge Constraint       & \third 0.0950 & \best 0.8824 & \best 0.6600 & \third 0.7997 & \second 0.7228 & \second 0.4758 & 0.6280 \\
            \midrule
            \multirow{1}{*}{Data} 
                & MEAD                      & 0.2125 & 0.7647 & 0.4700 & 0.7047 & 0.6187 & 0.4235 & 0.5735 \\
            \midrule
            \multirow{1}{*}{Ours} 
                & Full Setting              & \best 0.0550 & \third 0.8627 & \third 0.6000 & \second 0.8078 & \best 0.7305 & \third 0.4723 & \third 0.6522 \\
            
            \bottomrule
        \end{tabular}
    }
    \label{tab:ablation_table}
\end{table*}
\subsection{Qualitative Comparison}
We qualitatively compare \ourmethod{} with both general editing models and linear control baselines, as illustrated in Figure~\ref{fig:quality_general} and Figure~\ref{fig:quality_linear}.

\noindent\textbf{Comparison with General Editing Models.}
As shown in Figure~\ref{fig:quality_general}, existing general editing models struggle to simultaneously achieve clear expression editing and strong identity preservation. 
Several models, including \nbp, \qwen, \seed, and \longcat, preserve identity well but produce only weak expression changes, often resulting in barely noticeable edits. 
In contrast, \gpt\ generates more visible expression differences but introduces moderate identity drift. 
\flux\ performs the worst in both aspects, showing weak expression editing while severely degrading identity consistency. 
Compared with these methods, \ourmethod{} produces clear and recognizable expression changes while maintaining stable facial identity, achieving the best balance between semantic editing and identity preservation.

\noindent\textbf{Comparison with Linear Control Models.}
Figure~\ref{fig:quality_linear} compares continuous expression control across different methods. 
An ideal method should produce expression intensity that increases monotonically with the control parameter while preserving identity consistency. 
We first analyze the relatively simple expression \textit{Happy}. 
ConceptSlider and AttributeControl show limited linear response but quickly degrade identity as editing strength increases. 
SliderEdit exhibits a step-like behavior: expressions remain nearly unchanged for most control values and suddenly increase at higher strengths, accompanied by significant identity degradation. 
\kslider\ shows unstable behavior, where expression changes have little correlation with the control parameter. 
When moving to the more challenging expression \textit{Surprised}, the linear response of these methods further deteriorates and identity preservation becomes worse. 
In contrast, \ourmethod{} maintains a stable monotonic increase in expression intensity while preserving identity across the entire control range. 
Even for more difficult expressions such as \textit{Disgust}, our method continues to produce clear and controllable expression changes.

\begin{figure}[hbp]
    \centering
    \includegraphics[width=\linewidth]{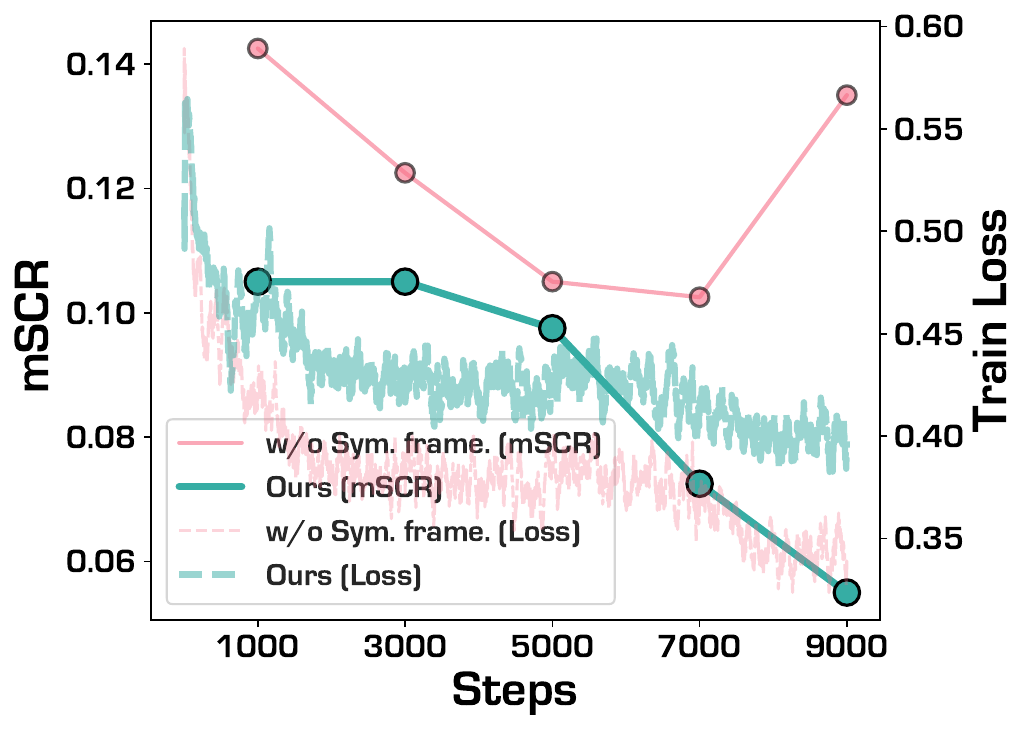}
    \caption{\textbf{Training dynamics of symmetric contrastive learning.} 
    The asymmetric variant reduces loss faster in early training but leads to higher structural confusion, while the symmetric framework achieves lower and more stable mSCR.}
    \label{fig:ablation_linechart}
\end{figure}

\subsection{Ablation Study}
\label{sec:ablation_study}

To validate the necessity of each component in \ourmethod{}, we conduct comprehensive ablation experiments, with quantitative results summarized in Table~\ref{tab:ablation_table}. Overall, the results reveal an inherent trade-off between expression editing capability and identity preservation: stronger editing often leads to identity degradation, while excessive identity constraints suppress effective expression transfer.

\noindent\textbf{Ablation on Loss Framework.}
We first analyze the roles of the identity loss and the contrastive loss. 
Removing the identity loss improves expression editing and disentanglement but significantly degrades identity consistency. 
The model tends to modify facial attributes such as hairstyle or skin texture to match the target expression, especially at large editing intensities, leading to clear identity drift and inconsistent facial appearance across edits. 
As illustrated in Fig.~\ref{fig:ablation_idloss}, the full model maintains stable identity while the variant without ID loss shows noticeable identity changes, confirming the importance of identity supervision for preserving subject consistency.

Conversely, removing the contrastive loss yields the highest identity similarity but leads to the weakest editing accuracy and the highest structural confusion. 
Without the contrastive objective, the model collapses toward reconstructing the source image instead of performing meaningful expression edits. 
As shown in Fig.~\ref{fig:ablation_scloss}, the model without contrastive supervision fails to separate semantically similar expressions, resulting in severe expression confusion. 
These results demonstrate that the two losses play complementary roles: identity loss stabilizes facial identity, while contrastive loss enhances expression disentanglement.

\noindent\textbf{Ablation on Symmetric Framework.}
We further compare the proposed symmetric training design with an asymmetric variant that applies contrastive supervision to only one branch. 
As shown in Fig.~\ref{fig:ablation_scloss}, removing the symmetric structure again leads to noticeable expression confusion. 
From the training dynamics in Fig.~\ref{fig:ablation_linechart}, the asymmetric model shows faster initial loss reduction but converges to worse solutions with lower editing accuracy and higher confusion rates. 
In contrast, the symmetric design acts as a structural regularizer: although it slows early convergence, the bidirectional constraints stabilize optimization and lead to better disentangled representations.

\noindent\textbf{Ablation on Triplet Formulations.}
We also compare three triplet formulations: \textit{Log-ratio}, \textit{Hinge}, and \textit{InfoNCE}. 
Log-ratio favors identity preservation but weakens expression editing, while Hinge maximizes editing strength at the cost of identity consistency. 
InfoNCE achieves the best balance between expression disentanglement and identity fidelity, and is therefore adopted as the default formulation.

\noindent\textbf{Ablation on Dataset.}
Finally, we evaluate the impact of training data by training the same architecture on the widely used MEAD dataset~\cite{wang2020mead}, with preprocessing details provided in Appendix~\ref{app:appendix_mead}. 
The MEAD-trained model consistently underperforms our full model across all metrics. 
This gap is mainly due to MEAD's limited identity diversity and discrete intensity annotations, which restrict fine-grained expression modeling and semantic disentanglement. 
In contrast, our \ourdataset{} provides richer identity variation and continuous soft-label supervision, enabling more precise and robust expression editing in the wild.

\subsection{User Study}
\begin{figure}[htbp]
    \centering
    \includegraphics[width=0.8\linewidth]{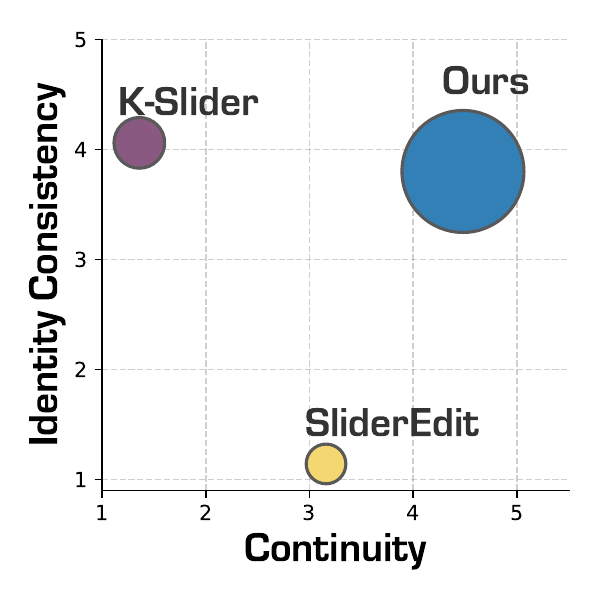}
    \caption{\textbf{User study results.} We show the trade-off between identity preservation and continuity of editing, annotated by human annotators. The size of the points indicates the HES scores of human annotators. }
    \label{fig:user_study}
\end{figure}

We conducted a user study with 2,400 images and 10 trained annotators who ranked three continuous editing methods on expression continuity and identity consistency. 
Mean scores (continuity, identity) are: \ourmethod{} (4.48, 3.80); \kslider{} (1.36, 4.06); SliderEdit (3.16, 1.14). 
As illustrated in Figure~\ref{fig:user_study}, human judgments are consistent with the machine-based evaluation. 
Overall, \ourmethod{} achieves the best balance, attaining the highest continuity while maintaining strong identity preservation.

\subsection{Expression Blend}
Human facial behavior often involves compound expressions~\cite{du2014compound,du2015compound,pan2023renderme}. 
To explore whether such compositionality emerges in the learned representation, we perform pairwise linear interpolation among six basic expressions, producing 15 zero-shot combinations. 
As shown in Fig.~\ref{fig:expression_blend} in the Appendix, several pairs generate perceptually coherent compound expressions, suggesting that the learned emotion manifold is continuous and compositional. 
However, some combinations collapse into a single dominant expression (e.g., Fear+Surprise) or produce unstable results due to physiological conflicts (e.g., Angry+Happy). 
Overall, 9 out of 15 combinations form plausible compound expressions, indicating that the learned representation supports linear composition while respecting implicit facial constraints and capturing meaningful compositional structure.

\section{Conclusion}

In this paper, we present \ourmethod{}, a framework for addressing semantic entanglement in facial expression editing. By shifting from discrete supervision to the continuous expression manifold defined by \ourdataset{} and evaluated through \ourbenchmark{}, our approach enables precise and linearly controllable editing via symmetric joint training. 
Extensive experiments demonstrate effectiveness of \ourmethod{} in four dimensions: structural confusion, expression accuracy, linear controllability, and identity preservation. 
Overall, this work establishes a standardized framework for fine-grained facial expression editing and advances research toward continuous and compositional facial affect manipulation.

\section*{Ethics Statement}
\noindent
All data in \ourdataset\ is collected from publicly available sources and used in compliance with their respective licenses and terms of use. The real-world subset is derived from existing public datasets (e.g., Human Images Dataset~\cite{snmahsa_human_images_dataset} and Matting Human Dataset~\cite{matting_human_datasets}, both distributed under the MIT License), while the anime subset consists of stylized fictional characters from publicly available media. We do not collect or use any private or login-restricted data.
Facial expression editing is a dual-use technology that may pose risks, such as misuse in identity-related scenarios. Our work focuses on expression manipulation and is intended for non-commercial academic research. We do not aim to alter identity or enable deceptive applications.
To mitigate potential risks, no personal metadata is retained, and the dataset is curated to exclude offensive content. We encourage responsible use of the dataset and models in compliance with applicable laws, regulations, and ethical guidelines.


\vspace{8pt}
{
    \small
    \bibliographystyle{ieeenat_fullname}
    \bibliography{main}

@String(IJCV = {Int. J. Comput. Vis.})

@String(CVPR= {IEEE Conf. Comput. Vis. Pattern Recog.})

@String(ICCV= {Int. Conf. Comput. Vis.})

@String(ECCV= {Eur. Conf. Comput. Vis.})

@String(TOG= {ACM Trans. Graph.})

@String(ICLR = {Int. Conf. Learn. Represent.})

@String(AAAI = {AAAI})

@String(IJCV  = {IJCV})

@String(CVPR  = {CVPR})

@String(ICCV  = {ICCV})

@String(ECCV  = {ECCV})

@String(TOG   = {ACM TOG})

@String(ICLR  = {ICLR})

@article{liu2025step1x,
  title={Step1x-edit: A practical framework for general image editing},
  author={Liu, Shiyu and Han, Yucheng and Xing, Peng and Yin, Fukun and Wang, Rui and Cheng, Wei and Liao, Jiaqi and Wang, Yingming and Fu, Honghao and Han, Chunrui and others},
  journal={arXiv preprint arXiv:2504.17761},
  year={2025}
}

@article{xu2025withanyone,
  title={Withanyone: Towards controllable and id consistent image generation},
  author={Xu, Hengyuan and Cheng, Wei and Xing, Peng and Fang, Yixiao and Wu, Shuhan and Wang, Rui and Zeng, Xianfang and Jiang, Daxin and Yu, Gang and Ma, Xingjun and others},
  journal={arXiv preprint arXiv:2510.14975},
  year={2025}
}

@article{wu2025qwenimagetechnicalreport,
  title={Qwen-image technical report},
  author={Wu, Chenfei and Li, Jiahao and Zhou, Jingren and Lin, Junyang and Gao, Kaiyuan and Yan, Kun and Yin, Sheng-ming and Bai, Shuai and Xu, Xiao and Chen, Yilei and others},
  journal={arXiv preprint arXiv:2508.02324},
  year={2025}
}

@article{guo2024pulid,
  title={Pulid: Pure and lightning id customization via contrastive alignment},
  author={Guo, Zinan and Wu, Yanze and Zhuowei, Chen and Zhang, Peng and He, Qian and others},
  journal={NeurIPS},
  year={2024}
}

@inproceedings{jiang2025infiniteyou,
  title={Infiniteyou: Flexible photo recrafting while preserving your identity},
  author={Jiang, Liming and Yan, Qing and Jia, Yumin and Liu, Zichuan and Kang, Hao and Lu, Xin},
  booktitle={ICCV},
  year={2025}
}

@article{zarei2025slideredit,
  title={SliderEdit: Continuous Image Editing with Fine-Grained Instruction Control},
  author={Zarei, Arman and Basu, Samyadeep and Pournemat, Mobina and Nag, Sayan and Rossi, Ryan and Feizi, Soheil},
  journal={arXiv preprint arXiv:2511.09715},
  year={2025}
}

@article{parihar2025kontinuous,
  title={Kontinuous Kontext: Continuous Strength Control for Instruction-based Image Editing},
  author={Parihar, Rishubh and Patashnik, Or and Ostashev, Daniil and Babu, R Venkatesh and Cohen-Or, Daniel and Wang, Kuan-Chieh},
  journal={arXiv preprint arXiv:2510.08532},
  year={2025}
}

@inproceedings{gandikota2024concept,
  title={Concept sliders: Lora adaptors for precise control in diffusion models},
  author={Gandikota, Rohit and Materzy{\'n}ska, Joanna and Zhou, Tingrui and Torralba, Antonio and Bau, David},
  booktitle={ECCV},
  year={2024}
}

@inproceedings{baumann2025continuous,
  title={Continuous, subject-specific attribute control in t2i models by identifying semantic directions},
  author={Baumann, Stefan Andreas and Krause, Felix and Neumayr, Michael and Stracke, Nick and Sevi, Melvin and Hu, Vincent Tao and Ommer, Bj{\"o}rn},
  booktitle={CVPR},
  year={2025}
}

@inproceedings{guerrero2024texsliders,
  title={Texsliders: Diffusion-based texture editing in clip space},
  author={Guerrero-Viu, Julia and Hasan, Milos and Roullier, Arthur and Harikumar, Midhun and Hu, Yiwei and Guerrero, Paul and Gutierrez, Diego and Masia, Belen and Deschaintre, Valentin},
  booktitle={SIGGRAPH},
  year={2024}
}

@inproceedings{sridhar2024prompt,
  title={Prompt sliders for fine-grained control, editing and erasing of concepts in diffusion models},
  author={Sridhar, Deepak and Vasconcelos, Nuno},
  booktitle={ECCV},
  year={2024}
}

@article{chang2025oneig,
  title={Oneig-bench: Omni-dimensional nuanced evaluation for image generation},
  author={Chang, Jingjing and Fang, Yixiao and Xing, Peng and Wu, Shuhan and Cheng, Wei and Wang, Rui and Zeng, Xianfang and Yu, Gang and Chen, Hai-Bao},
  journal={arXiv preprint arXiv:2506.07977},
  year={2025}
}

@inproceedings{peebles2023scalable,
  title={Scalable diffusion models with transformers},
  author={Peebles, William and Xie, Saining},
  booktitle={ICCV},
  year={2023}
}

@article{garibi2025tokenverse,
  title={Tokenverse: Versatile multi-concept personalization in token modulation space},
  author={Garibi, Daniel and Yadin, Shahar and Paiss, Roni and Tov, Omer and Zada, Shiran and Ephrat, Ariel and Michaeli, Tomer and Mosseri, Inbar and Dekel, Tali},
  journal={TOG},
  year={2025}
}

@article{zhuang2025vistorybench,
  title={Vistorybench: Comprehensive benchmark suite for story visualization},
  author={Zhuang, Cailin and Huang, Ailin and Hu, Yaoqi and Wu, Jingwei and Cheng, Wei and Liao, Jiaqi and Wang, Hongyuan and Liao, Xinyao and Cai, Weiwei and Xu, Hengyuan and others},
  journal={arXiv preprint arXiv:2505.24862},
  year={2025}
}

@article{zhong2025mod,
  title={Mod-adapter: Tuning-free and versatile multi-concept personalization via modulation adapter},
  author={Zhong, Weizhi and Yang, Huan and Liu, Zheng and He, Huiguo and He, Zijian and Niu, Xuesong and Zhang, Di and Li, Guanbin},
  journal={arXiv preprint arXiv:2505.18612},
  year={2025}
}

@article{kamenetsky2025saedit,
  title={SAEdit: Token-level control for continuous image editing via Sparse AutoEncoder},
  author={Kamenetsky, Ronen and Dorfman, Sara and Garibi, Daniel and Paiss, Roni and Patashnik, Or and Cohen-Or, Daniel},
  journal={arXiv preprint arXiv:2510.05081},
  year={2025}
}

@inproceedings{jain2025adaptivesliders,
  title={AdaptiveSliders: User-aligned Semantic Slider-based Editing of Text-to-Image Model Output},
  author={Jain, Rahul and Goel, Amit and Niinuma, Koichiro and Gupta, Aakar},
  booktitle={CHI},
  year={2025}
}

@article{cheng2022generalizable,
  title={Generalizable neural performer: Learning robust radiance fields for human novel view synthesis},
  author={Cheng, Wei and Xu, Su and Piao, Jingtan and Qian, Chen and Wu, Wayne and Lin, Kwan-Yee and Li, Hongsheng},
  journal={arXiv preprint arXiv:2204.11798},
  year={2022}
}

@inproceedings{zhu2022celebv,
  title={CelebV-HQ: A large-scale video facial attributes dataset},
  author={Zhu, Hao and Wu, Wayne and Zhu, Wentao and Jiang, Liming and Tang, Siwei and Zhang, Li and Liu, Ziwei and Loy, Chen Change},
  booktitle={ECCV},
  year={2022}
}

@misc{matting_human_datasets,
  title        = {Matting Human Datasets},
  author={aisegmentcn},
  howpublished = {\url{https://github.com/aisegmentcn/matting_human_datasets}},
  note         = {Accessed: 2026-03-03},
  year         = {2020}
}

@misc{snmahsa_human_images_dataset,
  title        = {Human Images Dataset (Men and Women)},
  author="snmahsa",
  howpublished = {\url{https://www.kaggle.com/datasets/snmahsa/human-images-dataset-men-and-women}},
  note         = {Accessed: 2026-03-03},
  year         = {2021}
}

@inproceedings{deng2019arcface,
    title={Arcface: Additive angular margin loss for deep face recognition},
    author={Deng, Jiankang and Guo, Jia and Xue, Niannan and Zafeiriou, Stefanos},
    booktitle={CVPR},
    year={2019}
    }

@inproceedings{schroff2015facenet,
  title={Facenet: A unified embedding for face recognition and clustering},
  author={Schroff, Florian and Kalenichenko, Dmitry and Philbin, James},
  booktitle={CVPR},
  year={2015}
}

@inproceedings{cheng2023dna,
  title={Dna-rendering: A diverse neural actor repository for high-fidelity human-centric rendering},
  author={Cheng, Wei and Chen, Ruixiang and Fan, Siming and Yin, Wanqi and Chen, Keyu and Cai, Zhongang and Wang, Jingbo and Gao, Yang and Yu, Zhengming and Lin, Zhengyu and others},
  booktitle={ICCV},
  year={2023}
}

@inproceedings{kim2022adaface,
  title={Adaface: Quality adaptive margin for face recognition},
  author={Kim, Minchul and Jain, Anil K and Liu, Xiaoming},
  booktitle={CVPR},
  year={2022}
}

@article{hu2022lora,
  title={Lora: Low-rank adaptation of large language models.},
  author={Hu, Edward J and Shen, Yelong and Wallis, Phillip and Allen-Zhu, Zeyuan and Li, Yuanzhi and Wang, Shean and Wang, Liang and Chen, Weizhu and others},
  journal={ICLR},
  year={2022}
}

@article{pan2023renderme,
  title={Renderme-360: A large digital asset library and benchmarks towards high-fidelity head avatars},
  author={Pan, Dongwei and Zhuo, Long and Piao, Jingtan and Luo, Huiwen and Cheng, Wei and Wang, Yuxin and Fan, Siming and Liu, Shengqi and Yang, Lei and Dai, Bo and others},
  journal={NeurIPS},
  year={2023}
}

@article{lipman2022flow,
  title={Flow matching for generative modeling},
  author={Lipman, Yaron and Chen, Ricky TQ and Ben-Hamu, Heli and Nickel, Maximilian and Le, Matt},
  journal={arXiv preprint arXiv:2210.02747},
  year={2022}
}

@inproceedings{radford2021learning,
  title={Learning transferable visual models from natural language supervision},
  author={Radford, Alec and Kim, Jong Wook and Hallacy, Chris and Ramesh, Aditya and Goh, Gabriel and Agarwal, Sandhini and Sastry, Girish and Askell, Amanda and Mishkin, Pamela and Clark, Jack and others},
  booktitle={ICLR},
  year={2021}
}

@inproceedings{oh2016deep,
  title={Deep metric learning via lifted structured feature embedding},
  author={Oh Song, Hyun and Xiang, Yu and Jegelka, Stefanie and Savarese, Silvio},
  booktitle={CVPR},
  year={2016}
}

@article{xu2025numerikontrol,
  title={NumeriKontrol: Adding Numeric Control to Diffusion Transformers for Instruction-based Image Editing},
  author={Xu, Zhenyu and Shen, Xiaoqi and Nan, Haotian and Zhang, Xinyu},
  journal={arXiv preprint arXiv:2511.23105},
  year={2025}
}

@article{wolf2026continuous,
  title={Continuous Control of Editing Models via Adaptive-Origin Guidance},
  author={Wolf, Alon and Katzir, Chen and Aberman, Kfir and Patashnik, Or},
  journal={arXiv preprint arXiv:2602.03826},
  year={2026}
}

@article{oord2018representation,
  title={Representation learning with contrastive predictive coding},
  author={Oord, Aaron van den and Li, Yazhe and Vinyals, Oriol},
  journal={arXiv preprint arXiv:1807.03748},
  year={2018}
}

@article{goodfellow2014generative,
  title={Generative adversarial nets},
  author={Goodfellow, Ian J and Pouget-Abadie, Jean and Mirza, Mehdi and Xu, Bing and Warde-Farley, David and Ozair, Sherjil and Courville, Aaron and Bengio, Yoshua},
  journal={NeurIPS},
  year={2014}
}

@inproceedings{ding2018exprgan,
  title={Exprgan: Facial expression editing with controllable expression intensity},
  author={Ding, Hui and Sricharan, Kumar and Chellappa, Rama},
  booktitle={AAAI},
  year={2018}
}

@inproceedings{pumarola2018ganimation,
  title={Ganimation: Anatomically-aware facial animation from a single image},
  author={Pumarola, Albert and Agudo, Antonio and Martinez, Aleix M and Sanfeliu, Alberto and Moreno-Noguer, Francesc},
  booktitle={ECCV},
  year={2018}
}

@inproceedings{liu2019stgan,
  title={Stgan: A unified selective transfer network for arbitrary image attribute editing},
  author={Liu, Ming and Ding, Yukang and Xia, Min and Liu, Xiao and Ding, Errui and Zuo, Wangmeng and Wen, Shilei},
  booktitle={CVPR},
  year={2019}
}

@inproceedings{choi2018stargan,
  title={Stargan: Unified generative adversarial networks for multi-domain image-to-image translation},
  author={Choi, Yunjey and Choi, Minje and Kim, Munyoung and Ha, Jung-Woo and Kim, Sunghun and Choo, Jaegul},
  booktitle={CVPR},
  year={2018}
}

@inproceedings{choi2020stargan,
  title={Stargan v2: Diverse image synthesis for multiple domains},
  author={Choi, Yunjey and Uh, Youngjung and Yoo, Jaejun and Ha, Jung-Woo},
  booktitle={CVPR},
  year={2020}
}

@inproceedings{karras2019style,
  title={A style-based generator architecture for generative adversarial networks},
  author={Karras, Tero and Laine, Samuli and Aila, Timo},
  booktitle={CVPR},
  year={2019}
}

@inproceedings{karras2020analyzing,
  title={Analyzing and improving the image quality of stylegan},
  author={Karras, Tero and Laine, Samuli and Aittala, Miika and Hellsten, Janne and Lehtinen, Jaakko and Aila, Timo},
  booktitle={CVPR},
  year={2020}
}

@inproceedings{shen2020interpreting,
  title     = {Interpreting the Latent Space of GANs for Semantic Face Editing},
  author    = {Shen, Yujun and Gu, Jinjin and Tang, Xiaoou and Zhou, Bolei},
  booktitle = {CVPR},
  year      = {2020}
}

@article{harkonen2020ganspace,
  title={Ganspace: Discovering interpretable gan controls},
  author={H{\"a}rk{\"o}nen, Erik and Hertzmann, Aaron and Lehtinen, Jaakko and Paris, Sylvain},
  journal={NeurIPS},
  year={2020}
}

@inproceedings{shen2021closed,
  title={Closed-form factorization of latent semantics in gans},
  author={Shen, Yujun and Zhou, Bolei},
  booktitle={CVPR},
  year={2021}
}

@inproceedings{yuksel2021latentclr,
  title={Latentclr: A contrastive learning approach for unsupervised discovery of interpretable directions},
  author={Y{\"u}ksel, O{\u{g}}uz Kaan and Simsar, Enis and Er, Ezgi G{\"u}lperi and Yanardag, Pinar},
  booktitle={ICCV},
  year={2021}
}

@inproceedings{ding2023diffusionrig,
  title={Diffusionrig: Learning personalized priors for facial appearance editing},
  author={Ding, Zheng and Zhang, Xuaner and Xia, Zhihao and Jebe, Lars and Tu, Zhuowen and Zhang, Xiuming},
  booktitle={CVPR},
  year={2023}
}

@inproceedings{jang2025controlface,
  title={Controlface: Harnessing facial parametric control for face rigging},
  author={Jang, Wooseok and Hong, Youngjun and Cha, Geonho and Kim, Seungryong},
  booktitle={CVPR},
  year={2025}
}

@article{wei2025magicface,
  title={Magicface: High-fidelity facial expression editing with action-unit control},
  author={Wei, Mengting and Varanka, Tuomas and Jiang, Xingxun and Khor, Huai-Qian and Zhao, Guoying},
  journal={arXiv preprint arXiv:2501.02260},
  year={2025}
}

@inproceedings{garau2021deca,
  title={DECA: Deep viewpoint-Equivariant human pose estimation using Capsule Autoencoders},
  author={Garau, Nicola and Bisagno, Niccolo and Br{\'o}dka, Piotr and Conci, Nicola},
  booktitle={ICCV},
  year={2021}
}

@inproceedings{danvevcek2022emoca,
  title={Emoca: Emotion driven monocular face capture and animation},
  author={Dan{\v{e}}{\v{c}}ek, Radek and Black, Michael J and Bolkart, Timo},
  booktitle={CVPR},
  year={2022}
}

@article{ho2020denoising,
  title={Denoising diffusion probabilistic models},
  author={Ho, Jonathan and Jain, Ajay and Abbeel, Pieter},
  journal={NeurIPS},
  year={2020}
}

@article{meng2021sdedit,
  title={Sdedit: Guided image synthesis and editing with stochastic differential equations},
  author={Meng, Chenlin and He, Yutong and Song, Yang and Song, Jiaming and Wu, Jiajun and Zhu, Jun-Yan and Ermon, Stefano},
  journal={arXiv preprint arXiv:2108.01073},
  year={2021}
}

@article{hertz2022prompt,
  title={Prompt-to-prompt image editing with cross attention control},
  author={Hertz, Amir and Mokady, Ron and Tenenbaum, Jay and Aberman, Kfir and Pritch, Yael and Cohen-Or, Daniel},
  journal={arXiv preprint arXiv:2208.01626},
  year={2022}
}

@inproceedings{brooks2023instructpix2pix,
  title={Instructpix2pix: Learning to follow image editing instructions},
  author={Brooks, Tim and Holynski, Aleksander and Efros, Alexei A},
  booktitle={CVPR},
  year={2023}
}

@inproceedings{zhang2023adding,
  title={Adding conditional control to text-to-image diffusion models},
  author={Zhang, Lvmin and Rao, Anyi and Agrawala, Maneesh},
  booktitle={ICCV},
  year={2023}
}

@misc{flux-2-2025,
    author={Black Forest Labs},
    title={{FLUX.2: Frontier Visual Intelligence}},
    year={2025},
    howpublished={\url{https://bfl.ai/blog/flux-2}},
}

@article{LongCat-Image,
      title={LongCat-Image Technical Report},
      author={Meituan LongCat Team and  Hanghang Ma and Haoxian Tan and Jiale Huang and Junqiang Wu and Jun-Yan He and Lishuai Gao and Songlin Xiao and Xiaoming Wei and Xiaoqi Ma and Xunliang Cai and Yayong Guan and Jie Hu},
        journal={arXiv preprint arXiv:2512.07584},
      year={2025}
}

@article{batifol2025flux,
  title={FLUX. 1 Kontext: Flow Matching for In-Context Image Generation and Editing in Latent Space},
  author={Labs, Black Forest and Batifol, Stephen and Blattmann, Andreas and Boesel, Frederic and Consul, Saksham and Diagne, Cyril and Dockhorn, Tim and English, Jack and English, Zion and Esser, Patrick and others},
  journal={arXiv preprint arXiv:2506.15742},
  year={2025}
}

@article{langner2010presentation,
  title={Presentation and validation of the Radboud Faces Database},
  author={Langner, Oliver and Dotsch, Ron and Bijlstra, Gijsbert and Wigboldus, Daniel HJ and Hawk, Skyler T and Van Knippenberg, AD},
  journal={Cognition and Emotion},
  year={2010}
}

@article{lundqvist1998karolinska,
  title={Karolinska directed emotional faces},
  author={Lundqvist, Daniel and Flykt, Anders and {\"O}hman, Arne},
  journal={Cognition and Emotion},
  year={1998}
}

@inproceedings{lucey2010extended,
  title={The extended cohn-kanade dataset (ck+): A complete dataset for action unit and emotion-specified expression},
  author={Lucey, Patrick and Cohn, Jeffrey F and Kanade, Takeo and Saragih, Jason and Ambadar, Zara and Matthews, Iain},
  booktitle={2010 ieee computer society conference on computer vision and pattern recognition-workshops},
  year={2010}
}

@inproceedings{yin20063d,
  title={A 3D facial expression database for facial behavior research},
  author={Yin, Lijun and Wei, Xiaozhou and Sun, Yi and Wang, Jun and Rosato, Matthew J},
  booktitle={7th international conference on automatic face and gesture recognition (FGR06)},
  year={2006},
}

@article{mollahosseini2017affectnet,
  title={Affectnet: A database for facial expression, valence, and arousal computing in the wild},
  author={Mollahosseini, Ali and Hasani, Behzad and Mahoor, Mohammad H},
  journal={IEEE transactions on affective computing},
  year={2017},
  publisher={IEEE}
}

@inproceedings{li2017reliable,
  title={Reliable Crowdsourcing and Deep Locality-Preserving Learning for Expression Recognition in the Wild},
  author={Li, Shan and Deng, Weihong and Du, JunPing},
  booktitle={CVPR},
  year={2017}
}

@inproceedings{BarsoumICMI2016,
    title={Training Deep Networks for Facial Expression Recognition with Crowd-Sourced Label Distribution},
    author={Barsoum, Emad and Zhang, Cha and Canton Ferrer, Cristian and Zhang, Zhengyou},
    booktitle={ICMI},
    year={2016}
}

@article{zhang2018facial,
  title={From facial expression recognition to interpersonal relation prediction},
  author={Zhang, Zhanpeng and Luo, Ping and Loy, Chen Change and Tang, Xiaoou},
  journal={IJCV},
  year={2018}
}

@inproceedings{wang2020mead,
  title={Mead: A large-scale audio-visual dataset for emotional talking-face generation},
  author={Wang, Kaisiyuan and Wu, Qianyi and Song, Linsen and Yang, Zhuoqian and Wu, Wayne and Qian, Chen and He, Ran and Qiao, Yu and Loy, Chen Change},
  booktitle={ECCV},
  year={2020}
}

@article{nagrani2020voxceleb,
  title={Voxceleb: Large-scale speaker verification in the wild},
  author={Nagrani, Arsha and Chung, Joon Son and Xie, Weidi and Zisserman, Andrew},
  journal={Computer Speech \& Language},
  year={2020}
}

@article{qiu2025emovid,
  title={EmoVid: A Multimodal Emotion Video Dataset for Emotion-Centric Video Understanding and Generation},
  author={Qiu, Zongyang and Wang, Bingyuan and Chen, Xingbei and He, Yingqing and Wang, Zeyu},
  journal={arXiv preprint arXiv:2511.11002},
  year={2025}
}

@article{zhang2025videmo,
  title={VidEmo: Affective-Tree Reasoning for Emotion-Centric Video Foundation Models},
  author={Zhang, Zhicheng and Wang, Weicheng and Zhu, Yongjie and Qin, Wenyu and Wan, Pengfei and Zhang, Di and Yang, Jufeng},
  journal={arXiv preprint arXiv:2511.02712},
  year={2025}
}

@inproceedings{liu2025f,
  title={F-bench: Rethinking human preference evaluation metrics for benchmarking face generation, customization, and restoration},
  author={Liu, Lu and Duan, Huiyu and Hu, Qiang and Yang, Liu and Cai, Chunlei and Ye, Tianxiao and Liu, Huayu and Zhang, Xiaoyun and Zhai, Guangtao},
  booktitle={ICCV},
  year={2025}
}

@article{zhu2025seed,
  title={Seed: A benchmark dataset for sequential facial attribute editing with diffusion models},
  author={Zhu, Yule and Liu, Ping and Zheng, Zhedong and Liu, Wei},
  journal={arXiv preprint arXiv:2506.00562},
  year={2025}
}

@misc{danbooruclip,
  author = {OysterQAQ},
  title = {DanbooruCLIP},
  year = {2023},
  publisher = {Hugging Face},
  howpublished = {\url{https://huggingface.co/OysterQAQ/DanbooruCLIP}},
  note = {Accessed: 2023-05-18}
}

@article{du2014compound,
  title={Compound facial expressions of emotion},
  author={Du, Shichuan and Tao, Yong and Martinez, Aleix M},
  journal={PNAS},
  year={2014}
}

@article{du2015compound,
  title={Compound facial expressions of emotion: from basic research to clinical applications},
  author={Du, Shichuan and Martinez, Aleix M},
  journal={Dialogues in clinical neuroscience},
  year={2015}
}

@article{ekman1992argument,
  title={An argument for basic emotions},
  author={Ekman, Paul},
  journal={Cognition \& emotion},
  year={1992},
  publisher={Taylor \& Francis}
}

@misc{google_nano_banana_pro,
  title        = {Nano Banana Pro: High‑Fidelity AI Image Generation and Editing Model},
  author       = {{Google}},
  year         = {2025},
  howpublished = {\url{https://www.androidcentral.com/apps-software/ai/googles-nano-banana-pro-and-more}}, 
  note         = {Accessed: 2026‑03},
}

@misc{openai_gpt_image15,
  title       = {Introducing GPT‑Image‑1.5},
  author      = {{OpenAI}},
  year        = {2025},
  howpublished = {\url{https://openai.com/index/new-chatgpt-images-is-here/}},
  note        = {Accessed: 2026-03},
}

@misc{bytedance_seedream45,
  title        = {Seedream 4.5: Advanced AI Image Generation Model},
  author       = {{ByteDance}},
  year         = {2025},
  howpublished = {\url{https://seed.bytedance.com/en/seedream4_5}}, 
  note         = {Accessed: 2026‑03},
}

@article{ye2025all,
  title={All-in-One Slider for Attribute Manipulation in Diffusion Models},
  author={Ye, Weixin and Zhu, Hongguang and Wang, Wei and Liu, Yahui and Wang, Mengyu},
  journal={arXiv preprint arXiv:2508.19195},
  year={2025}
}

@article{yin2025instructattribute,
  title={InstructAttribute: Fine-grained Object Attributes editing with Instruction},
  author={Yin, Xingxi and Zhang, Jingfeng and Deng, Yue and Li, Zhi and Li, Yicheng and Zhang, Yin},
  journal={arXiv preprint arXiv:2505.00751},
  year={2025}
}

@article{dalva2024fluxspace,
  title={Fluxspace: Disentangled semantic editing in rectified flow transformers},
  author={Dalva, Yusuf and Venkatesh, Kavana and Yanardag, Pinar},
  journal={arXiv preprint arXiv:2412.09611},
  year={2024}
}

@inproceedings{gandikota2025sliderspace,
  title={Sliderspace: Decomposing the visual capabilities of diffusion models},
  author={Gandikota, Rohit and Wu, Zongze and Zhang, Richard and Bau, David and Shechtman, Eli and Kolkin, Nick},
  booktitle={ICCV},
  year={2025}
}

@article{yang2025controllable,
  title={Controllable-Continuous Color Editing in Diffusion Model via Color Mapping},
  author={Yang, Yuqi and Chang, Dongliang and Fang, Yuanchen and SonG, Yi-Zhe and Ma, Zhanyu and Guo, Jun},
  journal={arXiv preprint arXiv:2509.13756},
  year={2025}
}

@inproceedings{cheng2025marble,
  title={MARBLE: Material Recomposition and Blending in CLIP-Space},
  author={Cheng, Ta Ying and Sharma, Prafull and Boss, Mark and Jampani, Varun},
  booktitle={CVPR},
  year={2025}
}

@inproceedings{sharma2024alchemist,
  title={Alchemist: Parametric control of material properties with diffusion models},
  author={Sharma, Prafull and Jampani, Varun and Li, Yuanzhen and Jia, Xuhui and Lagun, Dmitry and Durand, Fredo and Freeman, Bill and Matthews, Mark},
  booktitle={CVPR},
  year={2024}
}

@inproceedings{wang2019symmetric,
  title={Symmetric cross entropy for robust learning with noisy labels},
  author={Wang, Yisen and Ma, Xingjun and Chen, Zaiyi and Luo, Yuan and Yi, Jinfeng and Bailey, James},
  booktitle={ICCV},
  year={2019}
}
}

\clearpage
\newpage
\appendix
\noindent
\textbf{\LARGE Appendix}
\vspace{8pt}
\begin{figure*}[htbp]
    \vspace{-5mm}
    \centering    \includegraphics[width=0.9\textwidth]{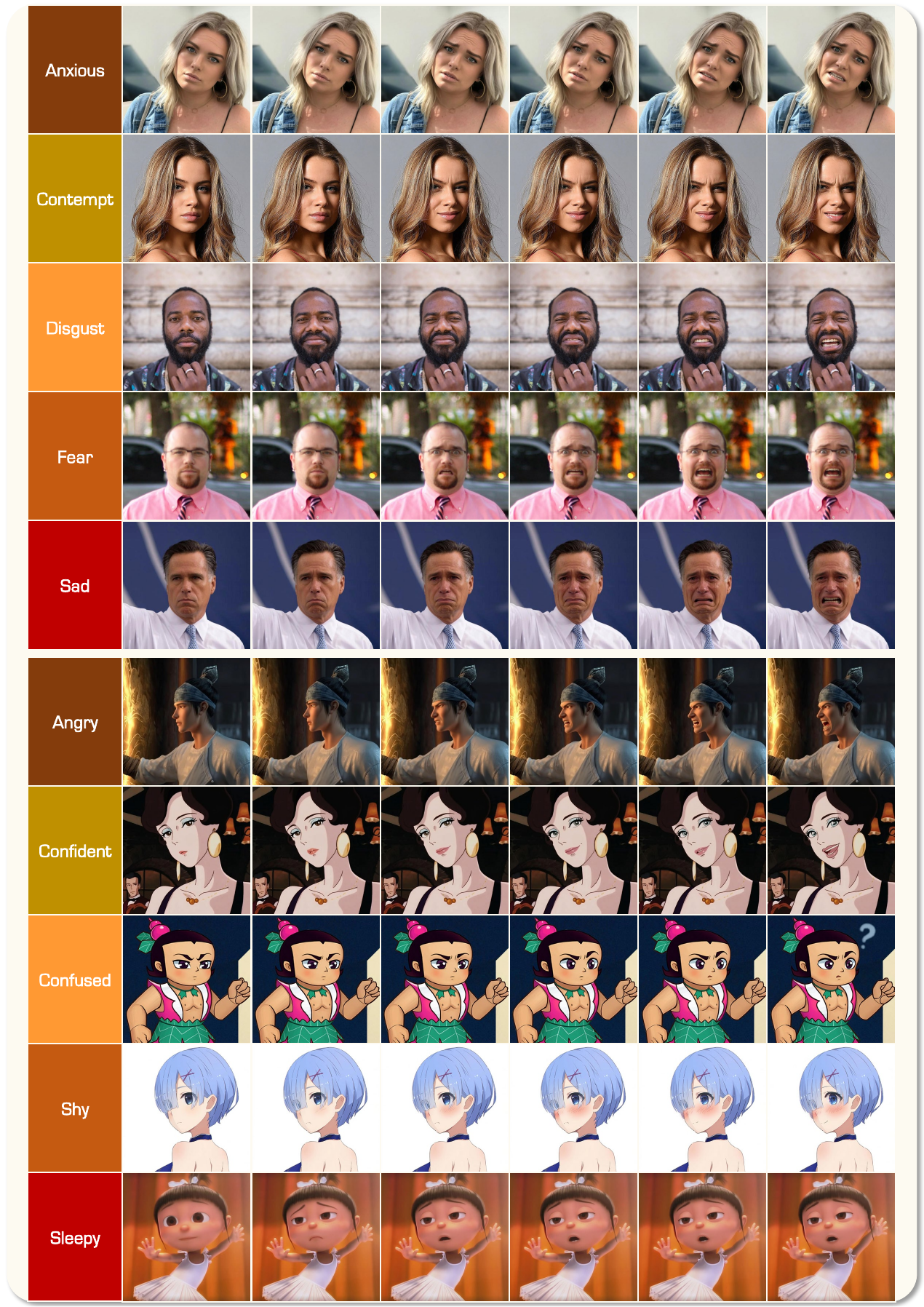}
    \caption{
    \textbf{Additional linear expression editing results.}
    We show the remaining ten expressions across both real and anime domains.
    The top row shows results on real images, while the bottom row shows results on anime images.
    Expression intensity increases from left to right for each expression.
    }
    \label{fig:quality_self}
\end{figure*}
\begin{figure*}[htbp]
    \centering
    \includegraphics[width=0.9\textwidth]{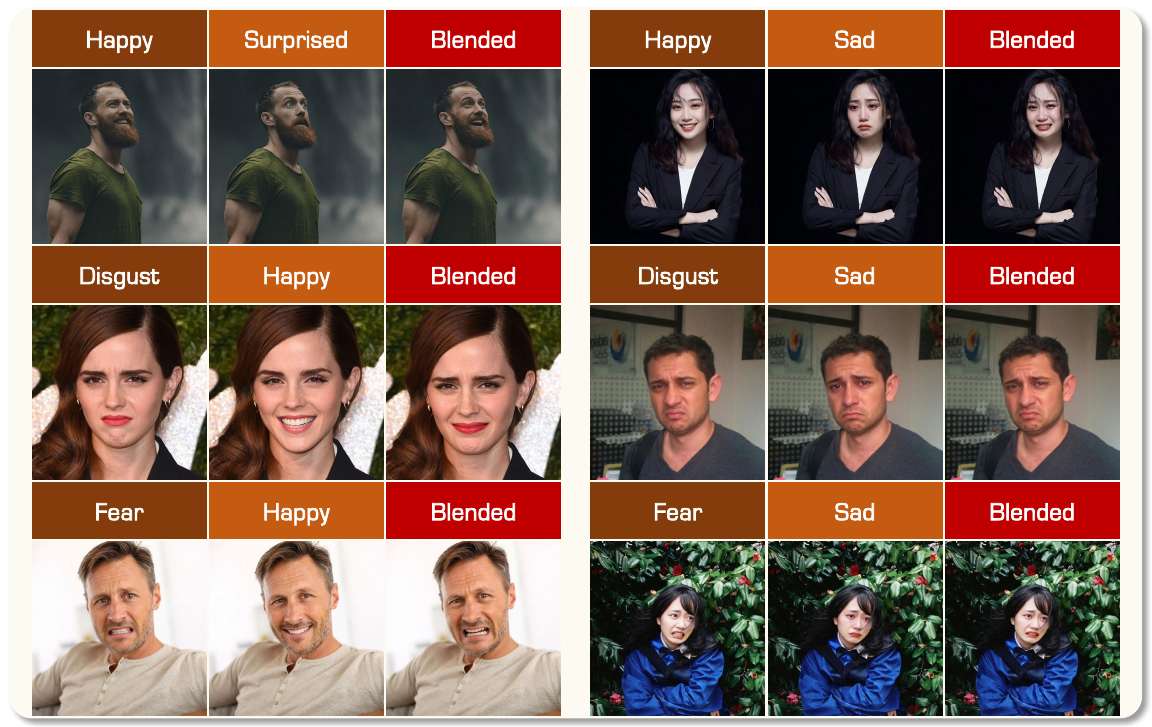}
    \caption{\textbf{Expression Blending Results.} Visualizing compositional facial expressions generated by smoothly blending multiple emotional categories in \ourmethod.}
    \label{fig:expression_blend}
\end{figure*}
\section{Details of the Symmetric Contrastive Loss}
\label{app:appendix_sc}














\subsection{Triplet Constraint Formulations}

In this section, we provide detailed formulations of the triplet constraint function 
$\mathcal{T}(G, P, N)$ used in Sec.~\ref{sec:method}. 
All features are extracted using a frozen CLIP image encoder to represent expression semantics 
and are $\ell_2$-normalized before distance computation. 
For brevity, we denote
$d_{G,P}=d(G,P)$ and $d_{G,N}=d(G,N)$ as cosine distances, and
$s_{G,P}=\mathrm{sim}(G,P)$ and $s_{G,N}=\mathrm{sim}(G,N)$ as cosine similarities.

\paragraph{Hinge-based Formulation.} 
The margin-based objective is

\begin{equation}
\mathcal{T}_{\mathrm{hinge}}(G,P,N) =
\max\bigl(
0,\;
d_{G,P} - d_{G,N} + m
\bigr),
\end{equation}

where $m$ is a fixed margin.

\paragraph{Log-Ratio Formulation.}

We adopt a smooth distance-ratio objective:

\begin{equation}
\mathcal{T}_{\mathrm{ratio}}(G,P,N) =
\log
\left(
\frac{d_{G,P} + \epsilon}
     {d_{G,N} + \epsilon}
\right),
\end{equation}

where $\epsilon$ is a small constant for numerical stability.

\paragraph{InfoNCE-style Formulation.}

The probabilistic contrastive objective is

\begin{equation}
\mathcal{T}_{\mathrm{nce}}(G,P,N)
=
- \log
\frac{\exp(s_{G,P}/\tau)}
{\sum_{x \in \{P,N\}} \exp(s_{G,x}/\tau)},
\end{equation}

where $\tau$ is a temperature parameter.

\subsection{Implementation Details}

Unless otherwise specified, we use the InfoNCE-style formulation 
with temperature $\tau=0.07$. 
For the hinge-based variant, the margin is set to $m=0.2$. 
For the log-ratio formulation, we set $\epsilon = 10^{-6}$ for numerical stability. 
All variants are evaluated under identical training schedules.
\section{Details of Experiment}
\label{app:appendix_exp}

To ensure reproducibility and clarity, we provide additional implementation details for \ourmethod{}. Training is conducted on 4 NVIDIA H200 GPUs.

\noindent\textbf{LoRA Configuration.}
We apply LoRA to major attention and MLP components of the diffusion transformer. 
Key hyperparameters are: rank = 64, $\alpha$ = 128, and dropout = 0.

\noindent\textbf{Training Hyperparameters.}
The models are optimized for 100 epochs using the AdamW optimizer with 
$\beta_1 = 0.9$, $\beta_2 = 0.999$, weight decay = 0.001, and $\epsilon = 1e{-8}$. 
The learning rate is set to $1e{-4}$ with cosine scheduling and 500 warmup steps. 
Mixed precision (bf16) is enabled to stabilize training. For the loss weights, we set $\lambda_{\mathrm{SC}} = 1.0$ 
(InfoNCE mode, symmetric) and $\lambda_{\mathrm{ID}} = 0.1$. 
The batch size per GPU is 4 with gradient accumulation steps = 1.
\section{Details of Dataset Ablation}
\label{app:appendix_mead}

\subsection{Dataset Overview}
Among human-centric dataset~\cite{cheng2023dna,pan2023renderme,cheng2022generalizable,wang2020mead,zhu2022celebv}, we choose MEAD~\cite{wang2020mead} to ablate on effectiveness of proposed dataset. The MEAD dataset \cite{wang2020mead} contains 7 discrete facial expressions captured from multi-view video sequences, with three intensity levels (low, medium, high). For our ablation, we only use the front-view subset and map its three intensity levels to continuous values 0.5,0.75,1.0 to match the input range of \ourmethod.
\subsection{Preprocessing and Triplet Construction}
Since MEAD provides video sequences, we uniformly sample frames to obtain independent images. From these sampled frames, we construct triplets ($ P_a, P_b, I_{orig}$ ) in the same manner as for \ourdataset~to train the symmetric contrastive framework. Each triplet consists of:

\begin{itemize}
\item $I_{orig}$ 1: the source frame.
\item $P_a, P_b$ 2: two frames of the same subject with distinct expressions.
\end{itemize}

Finally, we construct triplet data pairs from the same identities to conduct the symmetric contrastive training under our default configuration.
\section{Additional Qualitative Results}
\label{app:appendix_quality}

This section provides additional qualitative results for \ourmethod{}. 
We present more examples of linear expression editing across multiple expression categories, as well as additional expression blending results obtained through interpolation in the learned expression space.

\subsection{Additional Linear Expression Editing Results}

Figure~\ref{fig:quality_self} presents additional linear editing results for the remaining ten expressions across both real and anime domains. 
As the control parameter increases, the expression intensity changes smoothly while the facial identity remains consistent.

\subsection{Expression Blend Results}

Figure~\ref{fig:expression_blend} shows the examples of expression blending obtained through pairwise interpolation between basic expressions.
\section{Additional Dataset Details}
\label{app:appendix_dataset}

This section provides supplementary details of \ourdataset{}, including annotation and scoring prompts, together with additional dataset statistics.

\subsection{Annotation and Scoring Prompts}

We provide the prompt templates used in our annotation pipeline and expression scoring procedure. 
Table~\ref{tab:template_annotation_human} and Table~\ref{tab:template_annotation_anime} present the prompts used for statistical annotation of the human and anime subsets, respectively. 
These two templates are designed to extract structured semantic attributes for dataset analysis and are both based on \textbf{Qwen3-VL-235B-A22B}. 
Table~\ref{tab:template_score} shows the prompt used to assign expression intensity scores to images, which is based on \textbf{Gemini 3 Pro}.

\subsection{Dataset Statistics}
\begin{figure*}[htbp]
    \centering
    \begin{subfigure}{0.48\linewidth}
        \centering
        \includegraphics[width=\linewidth]{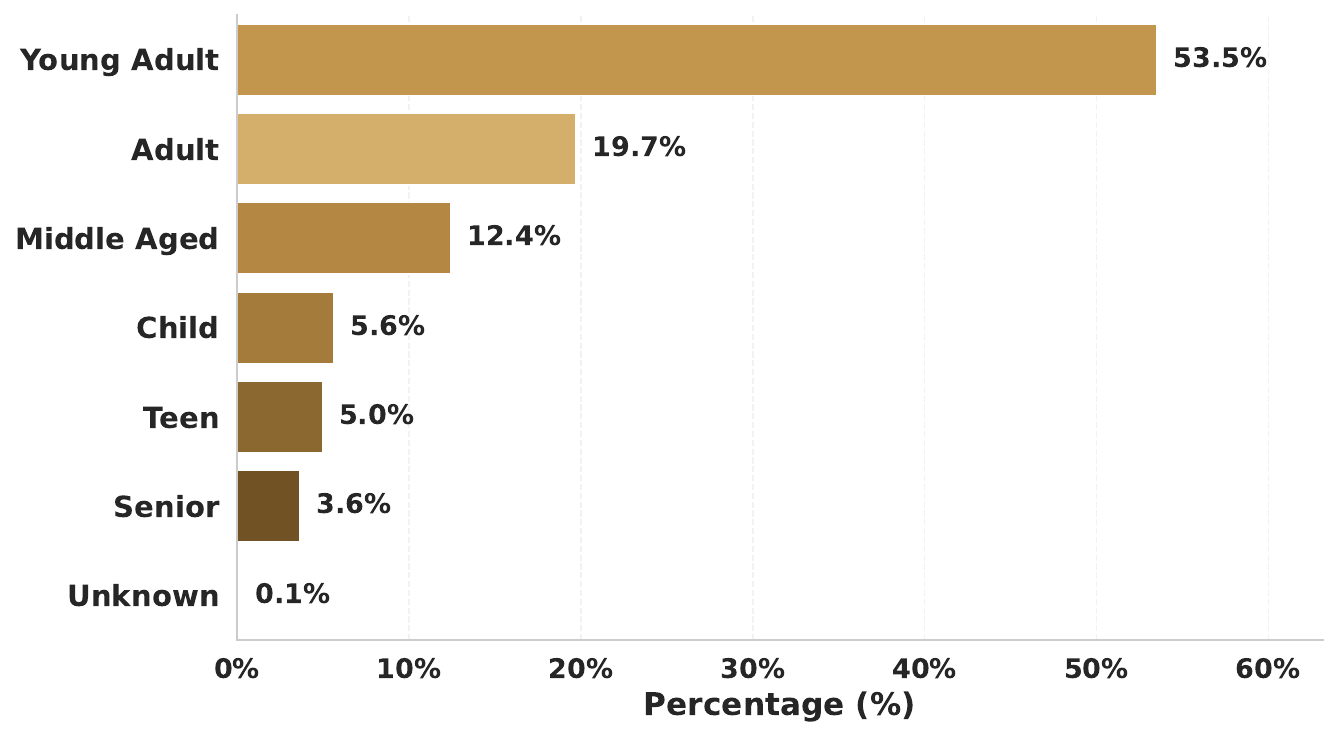}
        \caption{Age distribution in the real-world domain}
    \end{subfigure}
    \hfill
    \begin{subfigure}{0.48\linewidth}
        \centering
        \includegraphics[width=\linewidth]{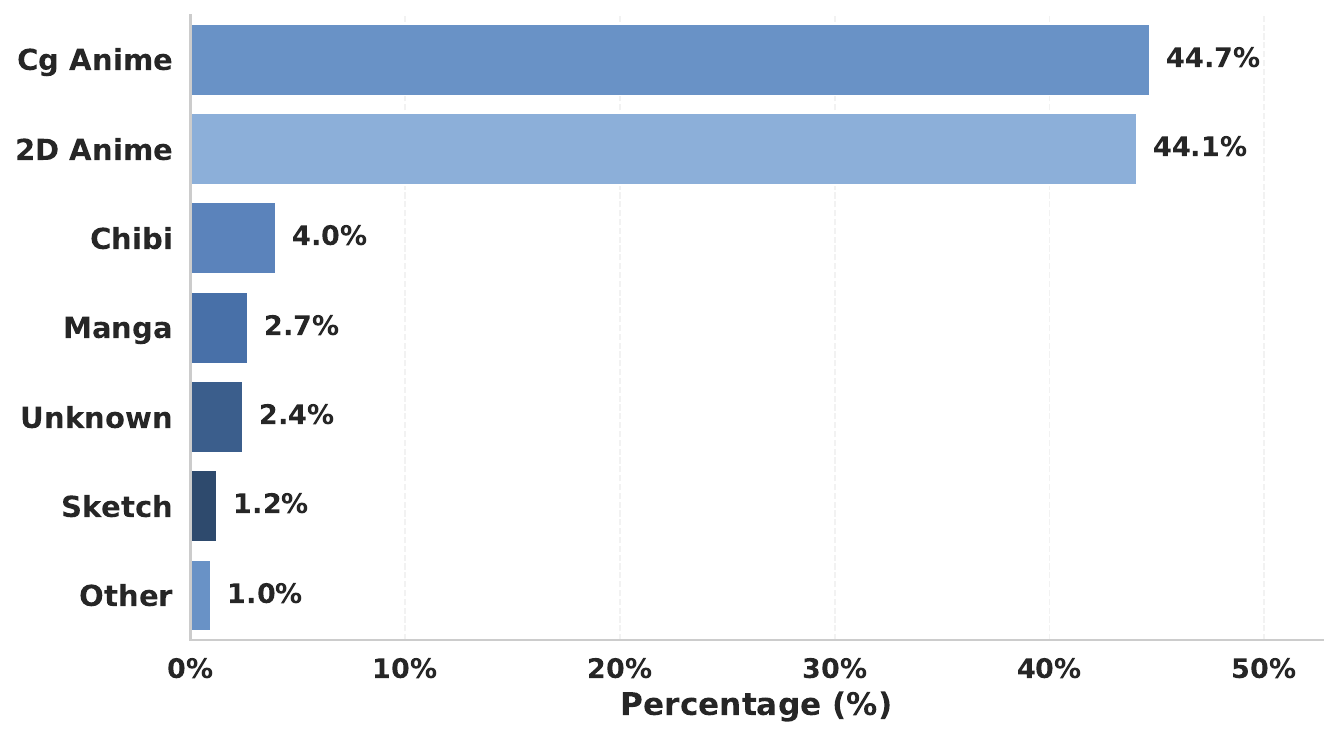}
        \caption{Style distribution in the anime domain}
    \end{subfigure}

    \caption{\textbf{Statistical distributions of annotated data in \ourdataset.}
    The results provide insights into the underlying data characteristics across real-world and anime domains.}
    
    \label{fig:dataset_distribution}
\end{figure*}

\begin{figure*}[t]
    \centering
    \begin{subfigure}{0.48\linewidth}
        \centering
        \includegraphics[width=\linewidth]{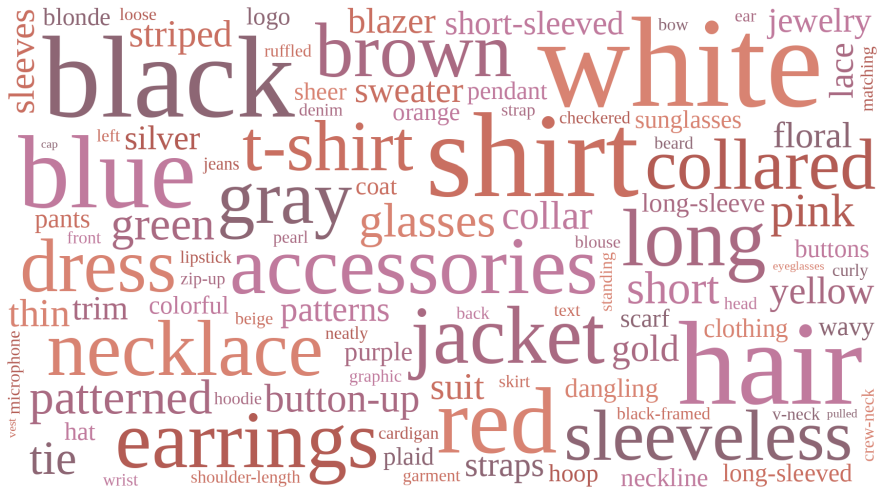}
        \caption{Real-world appearance descriptions}
    \end{subfigure}
    \hfill
    \begin{subfigure}{0.48\linewidth}
        \centering
        \includegraphics[width=\linewidth]{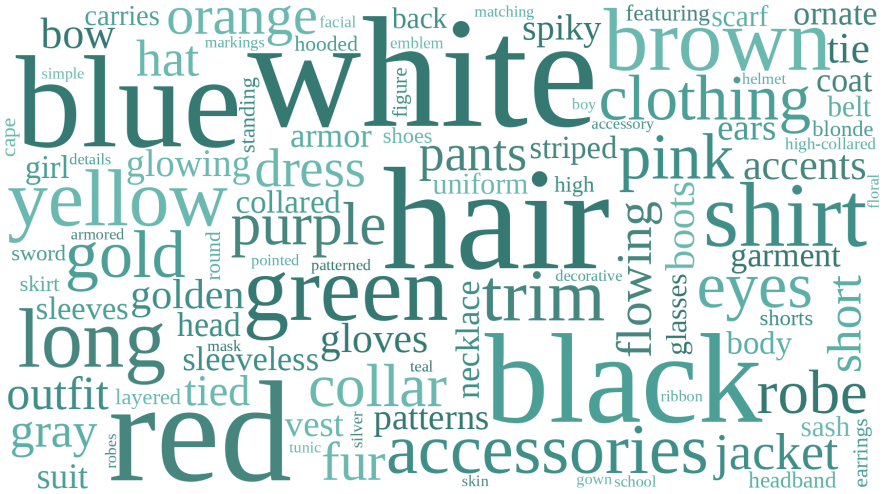}
        \caption{Anime-style appearance descriptions}
    \end{subfigure}

    \caption{\textbf{Visualization of appearance-related textual descriptions in \ourdataset.}
    The visualizations highlight the distribution and diversity of annotations across real-world and anime domains.}
    
    \label{fig:dataset_wordcloud}
\end{figure*}

We present the statistical analysis of \ourdataset{} in Fig.~\ref{fig:dataset_distribution} and Fig.~\ref{fig:dataset_wordcloud}, covering categorical distributions and textual description patterns across the real-world and anime domains.

From Fig.~\ref{fig:dataset_distribution}, the real-world subset is diverse but imbalanced, dominated by young adults (53.5\%), with children, teens, and seniors forming smaller proportions. Similar trends are observed in other attributes, where female samples are more frequent and light-to-medium skin tones constitute the majority, indicating that the dataset inherits non-uniform demographic characteristics. This bias reflects common patterns in portrait-centric internet images and introduces challenges for expression modeling. In contrast, the anime subset exhibits broader stylistic diversity, with CG and 2D anime each accounting for about 44\%, along with additional styles such as chibi, manga, and sketch. Compared with the real-world subset, the anime subset also shows a flatter age distribution. However, it contains more unknown labels in attributes like gender and age, suggesting that stylized characters are inherently more ambiguous under real-world categorization schemes, which increases the difficulty of consistent annotation and evaluation.

Fig.~\ref{fig:dataset_wordcloud} further reveals domain-specific textual patterns. The real-world subset emphasizes natural appearance cues such as clothing, hairstyle, and facial details, while the anime subset contains more stylized and visually distinctive descriptions. This difference highlights that expression editing in \ourdataset{} involves both visual transformation and domain-dependent semantic interpretation, requiring models to generalize across heterogeneous distributions. Overall, these statistics indicate that \ourdataset{} combines substantial diversity with realistic biases, making it a challenging and representative benchmark for fine-grained, diverse, and real-world facial expression editing.
\begin{table*}[htbp]
\centering
\caption{\small \textbf{Human Dataset Annotation Prompt Template.}}
\label{tab:template_annotation_human}

\begin{tcolorbox}[
    title=Human Dataset Annotation Prompt,
    colback={rgb,255:red,249;green,250;blue,255},
    colframe={rgb,255:red,109;green,153;blue,255},
    fonttitle=\bfseries
]

\textbf{System Context}

You are an image annotation assistant. For a single person image, output strict JSON only.

\vspace{6pt}

\textbf{Requirements}

\begin{itemize}[leftmargin=*, itemsep=2pt, parsep=0pt]
\item Describe only visible facts; do not infer identity, story, or intent.
\item \texttt{categorical} values must be selected from the provided enums.
\item All three fields in \texttt{descriptions} must be present.
\item Write all description sentences in English, each 8--25 words.
\end{itemize}

\vspace{6pt}

\textbf{JSON Schema}

\begin{lstlisting}[basicstyle=\ttfamily\small,columns=fullflexible,breaklines=true,showstringspaces=false]
{
  "categorical": {
    "gender": "male/female/androgynous/unknown",
    "age_group": "child/teen/young_adult/adult/middle_aged/senior/unknown",
    "skin_tone": "very_light/light/medium/dark/very_dark/unknown",
    "expression": "neutral/happy/sad/angry/surprised/fear/disgust/other/unknown"
  },
  "descriptions": {
    "appearance_sentence": "One sentence about clothing and accessories.",
    "action_sentence": "One sentence about action or pose.",
    "background_sentence": "One sentence about scene/background."
  }
}
\end{lstlisting}

\end{tcolorbox}
\end{table*}

\begin{table*}[htbp]
\centering
\caption{\small \textbf{Anime Dataset Annotation Prompt Template.}}
\label{tab:template_annotation_anime}

\begin{tcolorbox}[
    title=Anime Dataset Annotation Prompt,
    colback={rgb,255:red,249;green,250;blue,255},
    colframe={rgb,255:red,109;green,153;blue,255},
    fonttitle=\bfseries
]

\textbf{System Context}

You are an anime image annotation assistant. For a single character image, output strict JSON only.

\vspace{6pt}

\textbf{Requirements}

\begin{itemize}[leftmargin=*, itemsep=2pt, parsep=0pt]
\item Describe only visible facts; do not infer identity, story, or intent.
\item \texttt{categorical} values must be selected from the provided enums.
\item All three fields in \texttt{descriptions} must be present.
\item Write all description sentences in English, each 8--25 words.
\end{itemize}

\vspace{6pt}

\textbf{JSON Schema}

\begin{lstlisting}[basicstyle=\ttfamily\small,columns=fullflexible,breaklines=true,showstringspaces=false]
{
  "categorical": {
    "gender": "male/female/androgynous/unknown",
    "age_group": "child/teen/young_adult/adult/middle_aged/senior/unknown",
    "expression": "neutral/happy/sad/angry/surprised/fear/disgust/other/unknown",
    "anime_style": "2d_anime/chibi/manga/sketch/cg_anime/other/unknown"
  },
  "descriptions": {
    "appearance_sentence": "One sentence about clothing and accessories.",
    "action_sentence": "One sentence about action or pose.",
    "background_sentence": "One sentence about scene/background."
  }
}
\end{lstlisting}

\end{tcolorbox}
\end{table*}

\begin{table*}[htbp]
\centering
\caption{\small \textbf{Facial Expression Scoring Prompt Template.} The same prompt is applied to both human and anime domains, highlighting the domain-agnostic nature of our scoring pipeline.}
\label{tab:template_score}

\begin{tcolorbox}[
    title=Facial Expression Scoring Prompt,
    colback={rgb,255:red,250;green,255;blue,245},
    colframe={rgb,255:red,76;green,175;blue,80},
    fonttitle=\bfseries
]

\textbf{Role and Task Definition}

You are an expert AI specialized in analyzing facial expressions in both photorealistic and anime-style images.
Your task is to analyze the input image and determine the intensity score for specific expression categories.

\vspace{6pt}

\textbf{Target Emotion Definitions}

Analyze the image based on the following visual definitions.
Crucially, pay close attention to the distinctions between similar expressions (e.g., Fear vs.\ Surprise, Anger vs.\ Disgust).

\begin{itemize}[leftmargin=*, itemsep=2pt, parsep=0pt]
\item Happiness: Faintly smiling, cheerful, ecstatic; corners of the mouth raised or laughing.
\item Sadness: Somber, sorrowful, devastated; frowning with downcast eyes or tears.
\item Anger: Annoyed, angry, furious; furrowed brows and glaring eyes.
\item Fear: Wary, frightened, terrified; knitted brows and pupil constriction.
\item Surprise: Taken aback, amazed, stunned; eyes wide open or jaw dropped.
\item Disgust: Distasteful, repulsed, gagging; wrinkling the nose or raising the upper lip.
\item Embarrassment: A shy smile, blushing cheeks, avoiding eye contact, looking down, or covering the face with hands.
\item Confidence: Self-assured smile, corners of the mouth raised, sharp and firm gaze, slightly smug.
\item Confusion: Knitted brows, looking aside or rolling the eyes, head tilted in deep thought.
\item Drowsiness: Heavy drooping eyelids, yawning, lethargic or low-energy look.
\item Contempt: Asymmetrical smirk (one corner raised), sneering, looking down on someone.
\item Nervousness: Visible sweat drops, tense facial muscles, uneasy or restless gaze.
\end{itemize}

\vspace{6pt}

\textbf{Output Format Requirements}

\begin{itemize}[leftmargin=*, itemsep=2pt, parsep=0pt]
\item Return format: Output a standard JSON object.
\item Keys: Use the exact emotion labels provided above.
\item Values: The intensity score (a float between 0.00 and 1.00).
\item Constraint: Do not explain your reasoning. Output only the JSON object. Do not include Markdown formatting such as \texttt{```json ... ```}.
\end{itemize}

\vspace{6pt}

\textbf{Output Example}

\begin{lstlisting}[basicstyle=\ttfamily\small,columns=fullflexible,breaklines=true,showstringspaces=false]
{
  "Happiness": 0.90,
  "Surprise": 0.75,
  "Anger": 0.05,
  ...
}
\end{lstlisting}

\end{tcolorbox}
\end{table*}

\end{document}